\documentclass{article}

\usepackage{arxiv}

\usepackage[utf8]{inputenc} 
\usepackage[T1]{fontenc}    
\usepackage{hyperref}       
\usepackage{url}            
\usepackage{booktabs}       
\usepackage{amsfonts}       
\usepackage{nicefrac}       
\usepackage{microtype}      
\usepackage{lipsum}
\usepackage{graphicx}

\usepackage{tabularx}
\usepackage{multirow}
\usepackage{rotating}
\usepackage{xcolor}
\usepackage{overpic} 
\usepackage{amsmath}
\usepackage{caption}

\usepackage[final]{changes}

\title{A quantitative analysis of knowledge-learning preferences in large language models in molecular science}

\author{
  Pengfei Liu \\
  School of Computer Science and Engineering, \\
  Sun Yat-sen University\\
  Peng Cheng Laboratory\\
  \And
  Jun Tao \\
  School of Computer Science and Engineering, \\
  Sun Yat-sen University\\
  \And
  Zhixiang Ren\thanks{Corresponding author} \\
  Peng Cheng Laboratory\\
  \texttt{jason.zhixiang.ren@outlook.com} \\
}

\begin{document}
\maketitle

\begin{abstract}

Deep learning has significantly advanced molecular modeling and design, enabling efficient understanding and discovery of novel molecules.
In particular, large language models (LLMs) introduce a fresh research paradigm to tackle scientific problems from a natural language processing (NLP) perspective.
\added{LLMs significantly enhance our understanding and generation of molecules, often surpassing existing methods with their capabilities to decode and synthesize complex molecular patterns.}
However, two key issues remain: how to quantify the match between model and data modalities and how to identify the knowledge-learning preferences of models.
To address these challenges, we propose a multi-modal benchmark, named ChEBI-20-MM, and perform 1263 experiments to assess the model's compatibility with data modalities and knowledge acquisition.
Through the modal transition probability matrix, we provide insights into the most suitable modalities for tasks.
Furthermore, we introduce a statistically interpretable approach to discover context-specific knowledge mapping by localized feature filtering.
\added{Our analysis offers an exploration of the learning mechanism and paves the way for advancing LLMs in molecular science.}
\end{abstract}


\section*{}

\label{sec:intro}
Molecular modeling and design are pivotal in the discovery and development of new molecules, serving a wide range of applications from drug discovery to materials science.
These fields are fundamentally reliant on the ability to not only discover novel molecules but also to understand their structures and properties comprehensively.
Such capabilities are crucial for developing targeted therapies in medicine, creating advanced materials with specific functionalities, and enhancing our fundamental understanding of chemical and biological processes.
By unlocking the potential of molecular insights, we pave the way for breakthroughs that can transform technologies and improve human health.
Nevertheless, traditional approaches to discovering new molecules or refining existing ones are often arduous, costly, and fraught with a high risk of failure.
In contrast, modern computational approaches significantly improve the creation and modification of molecules.
Yet, they require considerable computational resources.


\begin{figure*}[!htb]
\centering
\includegraphics[width=0.93\textwidth]{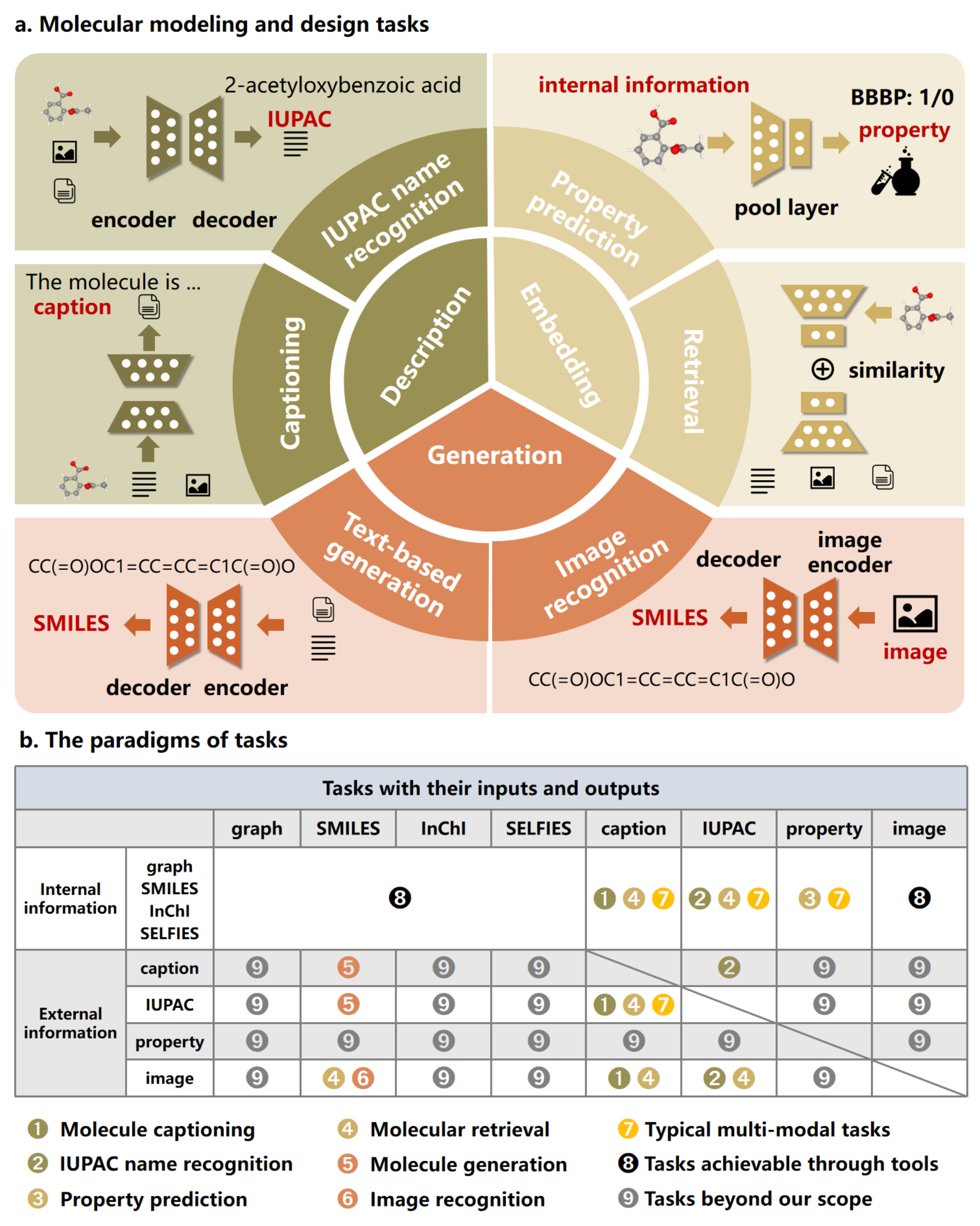}
\caption{\textbf{The paradigm of the analysis.}
\textbf{a. Molecular modeling and design tasks}, showcasing six task types with their standard modeling methods and data examples.
\textbf{b. The paradigms of tasks}, we divide common molecular data into two categories: internal and external information. Internal information, integral to molecular representation, can be converted through various tools. External information is more accessible to human understanding. Additionally, this part highlights the research scope of our analysis, detailing the input and output for each task.}
\label{fig:Conceptual-review}
\end{figure*}

Artificial Intelligence (AI) methods have made significant strides in this domain.
They offer faster computations and the ability to handle large datasets.
In recent years, AI-driven approaches such as Generative Adversarial Networks (GANs) \cite{creswell2018generative} and diffusion models \cite{xu2021geodiff} have been designed for molecule generation.
These techniques simulate the creative process of designing new molecular structures, thereby expanding the potential for groundbreaking discoveries in molecular science.
The task of understanding molecular structures and properties, particularly through molecular description and embedding, has been significantly advanced by machine learning techniques \cite{heikamp2014support} and graph models \cite{xu2018powerful}\cite{kipf2017semisupervised}\cite{veličković2018graph}.
These approaches have enhanced our ability to accurately represent and interpret the intricate details of molecules.
However, most AI models often face challenges in generalizability and flexibility, requiring extensive data for training to ensure accuracy and require specific adjustments for each molecular task.

Transformers \cite{vaswani2017attention} offer an advantage with their robust text encoding and generation capabilities.
These models can be fine-tuned with minimal task-specific adjustments, making them more versatile and efficient in molecular modeling and design. 
As depicted in Figure \ref{fig:Conceptual-review}.a, we organize six pivotal molecular tasks into three objectives, each addressing specific scientific needs.
In the \textbf{description} category, tasks such as captioning and IUPAC name recognition are pivotal for understanding complex molecular structures.
Molecular captioning renders intricate chemical details more comprehensible.
IUPAC name recognition is crucial for ensuring precise chemical identity across various platforms, despite the complexities of nomenclature rules and the scarcity of robust automated tools capable of handling such challenges.
In the \textbf{embedding} category, we focus on property prediction and retrieval.
These tasks are vital for extracting meaningful insights from molecular data and facilitating the prediction of molecular behaviors under various conditions.
The \textbf{generation} category encompasses text-based molecule generation and optical recognition, which aim to create or recreate accurate molecular representations.
Text-based generation enables the creation of novel molecules through predictive models, while optical recognition is essential for digitizing historical chemical data, ensuring that legacy research continues to inform and enhance contemporary scientific endeavors.

Moreover, since the advent of ChatGPT \cite{schulman2022chatgpt} and GPT-4 \cite{DBLP:journals/corr/abs-2303-08774}, LLMs have emerged as a groundbreaking trend, especially in molecular science.
LLMs, with their advanced capabilities in processing and generating human-like text, present a novel paradigm for understanding and designing molecular structures.
This paradigm is known as scientific language modeling (SLM).
Their ability to assimilate and analyze vast amounts of textual data can provide unprecedented insights, overcoming some of the limitations of traditional AI methods.
This new capability combines accuracy and novelty for improved outcomes, termed chemical knowledge.
Its effectiveness depends on factors like input data, model architecture, and training strategies.
However, current assessments of this capability remain incomplete.

Existing related work in molecular science, like the molecule-generation survey \cite{du2022molgensurvey}, lacks comprehensive model comparisons and has a limited task scope.
The knowledge-driven review \cite{fang2022knowledge} categorizes molecular learning but misses detailed method comparisons and dataset discussions.
And recent benchmarks, such as one testing ChatGPT \cite{guo2023indeed}, cover eight chemical tasks, each providing unique chemical insights.
Mol-Instructions \cite{fang2023mol} offers a dataset for fine-tuning with various molecule and protein instructions, boosting biomolecular understanding in LLMs.
Multi-modal data and methodologies are crucial in the molecular domain, as they naturally integrate diverse data types, providing complementary insights that enhance the understanding of molecular behaviors.
However, these papers lack multi-modal content and do not sufficiently explore the models' chemical knowledge.

In this study, we regard different representations of molecules as distinct modalities, transcending conventional formats like 1D text, 2D graphs, or images.
\added{We introduce ChEBI-20-MM, a comprehensive multi-modal benchmark encompassing 32,998 molecules.}
These molecules can be characterized by 1D textual chemical descriptors, including \textbf{SMILES} (Simplified Molecular Input Line Entry System) \cite{weininger1988smiles}, \textbf{InChI} (International Chemical Identifier) \cite{heller2013inchi}, and \textbf{SELFIES} (Self-Referencing Embedded Strings) \cite{krenn2020self}, along with their corresponding 2D graphs generated by cheminformatics tools such as RDKit \cite{landrum2013rdkit}.
These representations are classified as the molecules' internal information because they encapsulate the molecular essence.
Furthermore, these modalities can be interconverted using cheminformatics tools, although the processes of their conversion are not the focal point of our investigation.
Conversely, modalities such as molecular captions, \textbf{IUPAC} (International Union of Pure and Applied Chemistry) \cite{long1983limit} names, and images, which enhance human comprehension, are categorized as external information.
The intrinsic connections between different modalities, such as images, 2D graphs, and text, are profound.
For instance, while graphs can be directly transformed into images with minimal information loss, the reverse process from images to graphs tends to incur significant data loss.
Similarly, converting molecular internal information into captions or IUPAC names typically results in a reduction of informational content.

To address these losses, models are essential for bridging these gaps.
We construct a modal transition probability matrix to analyze the efficiency of modal conversion.
As depicted in Figure \ref{fig:Conceptual-review}.b, the classification is reflected in the evaluation framework of the six tasks (Task 1-6) and their respective modalities.
Our benchmark not only evaluates the performance variances across modalities but also can assess the multi-modal synergistic performance.
It includes the translation from various types of internal information to external information and generating captions based on IUPAC names, as represented by Task 7, which exemplifies typical multi-modal tasks.
It should be noted that our benchmark encompasses seven modalities.
Furthermore, to facilitate the evaluation of embedding ability, we integrate the MoleculeNet dataset \cite{wu2018moleculenet} and classify molecular properties as external information, treating the property prediction as a form of translation.

In summary, this study analyzes the LLMs in molecular modeling and design.
We categorize six common molecular tasks into three distinct objectives: description, embedding, and generation, as vividly depicted in Figure \ref{fig:Conceptual-review}.
Moreover, we establish a unified multi-modal benchmark ChEBI-20-MM and conduct experiments to assess the compatibility of data modalities, model architectures, and diverse task types, examining their impact on task performance.
Additionally, our end-to-end visual method showcases the discovery of modeling insights embedded with chemical knowledge.
Overall, our main contributions are the following:
\begin{itemize}
    \item This work quantitatively analyzes LLMs in molecule modeling, categorizes existing models, and presents a multi-modal benchmark (ChEBI-20-MM) for performance evaluation, backed by \textbf{1263} experiments.
    \item We explore the modal transition probability matrix and identify the \textbf{best match} within typical data modalities and model architectures.
    \item We propose a statistically interpretable approach to showcase the \textbf{knowledge acquisition} through localized feature filtering.
\end{itemize}



\section*{Results}
\label{sec:eva}
To obtain comprehensive insights into the factors of model performance, it is essential to conduct a systematic and unbiased comparison across various LLMs.
We introduce ChEBI-20-MM, a multi-modal benchmark developed from the ChEBI-20 dataset that integrates various data, including SMILES, InChI, IUPAC names, SELFIES, captions, and images.
This benchmark is designed to assess the chemical knowledge of models across tasks such as molecule generation, image and IUPAC recognition, molecular captioning, and retrieval, featuring in-depth analysis of the compatibility between modalities and models.
We explore the models' tendencies for knowledge acquisition through two typical tasks, employing a localized feature filtering method.
These processes demonstrate the feasibility of cross-domain knowledge-learning analysis.
Case studies further validate the utility of this analysis process.

\subsection*{Performance evaluation of tasks}
\label{sec: Experimental Results}

\begin{figure*}[t!]
\centering
\includegraphics[width=\textwidth]{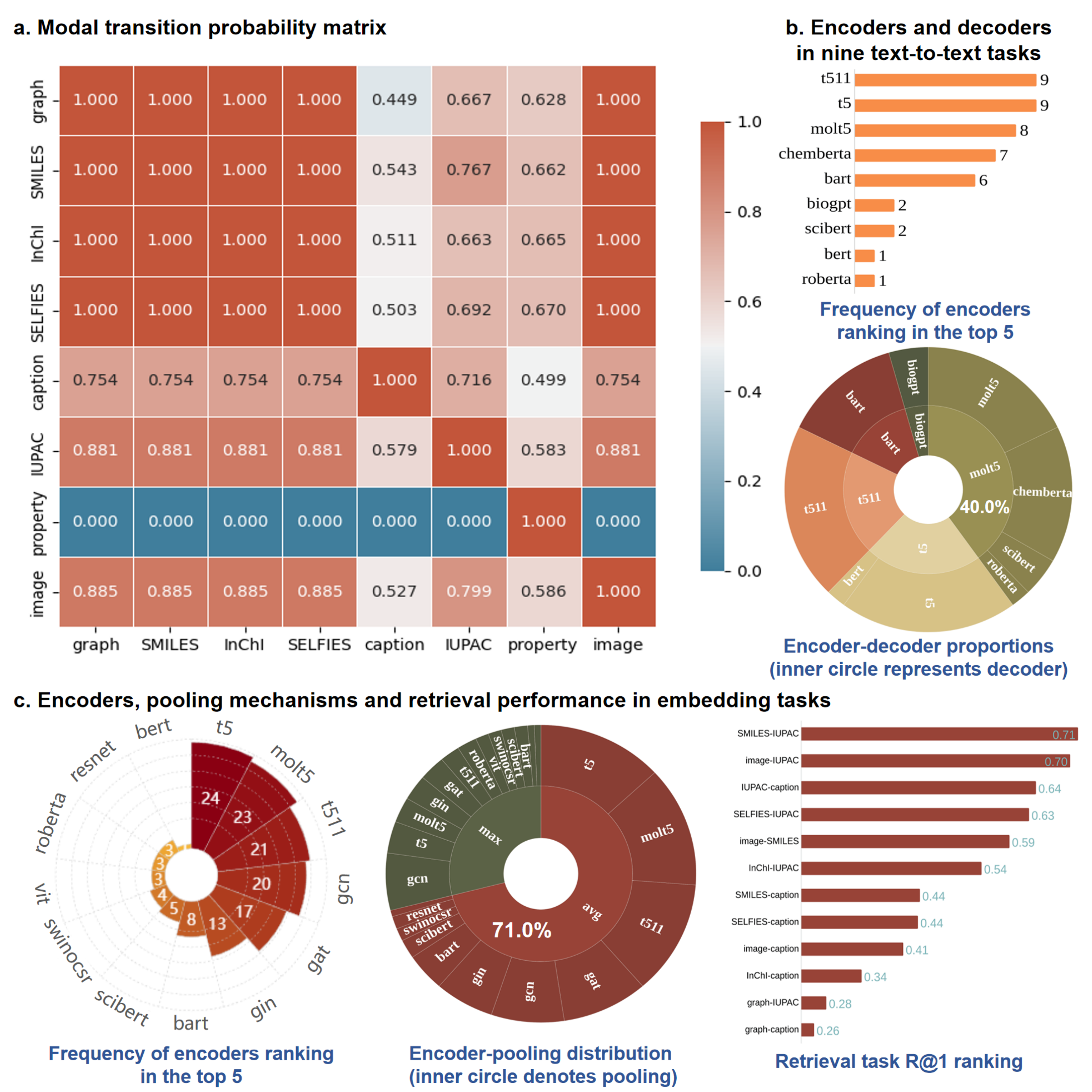}
\caption{\textbf{Results of benchmark.}
\textbf{a. Modal transition probability matrix.}
This matrix presents the performance in text generation and property prediction tasks. The vertical axis represents input modalities, while the horizontal axis denotes output modalities.
\textbf{b. Encoders and decoders in nine text-to-text tasks.}
This illustration highlights the frequency of various models appearing in the top 5 rankings.
The T5-based models exhibit a dominant presence.
\textbf{c. Encoders, pooling mechanisms, and retrieval performance in embedding tasks.}
Alongside model rankings, the figure indicates that average pooling is a preferred choice for the pooling layer.}
\label{fig:result}
\end{figure*}

For the analysis of modality adaptation to different tasks, we select representative metrics indicative of successful task completion, such as the METEOR score for molecule captioning and the area under the receiver operating characteristic curve (ROC\_AUC) for property classification tasks.
We fill out the modal transition probability matrix, which provides insights into the most suitable data modalities for different task types, as depicted in Figure \ref{fig:result}.a.
Regarding model analysis for different tasks, Figure \ref{fig:result}.b and Figure \ref{fig:result}.c demonstrate the frequency of appearances of encoders and decoders in nine text-to-text tasks, as well as eight encoders and two pooling mechanisms in embedding tasks. 
Detailed experimental results are shown in  Supplementary Section C. Table 3-Table 9.
We also incorporate typical traditional machine learning methods, graph models, and transformer-based models for property prediction task analysis in Tables 6 and 7.

\textbf{Modal transition probability matrix.}
In our study, we construct a modal transition probability matrix based on outcomes from various text generation tasks, such as any-to-text, encompassing 13 distinct tasks derived from the ChEBI-20-MM dataset and experimental settings.
Additionally, results for molecular property prediction tasks are obtained from MoleculeNet and mapped to the paradigms illustrated in Figure \ref{fig:Conceptual-review}.b.
This matrix features the inputs along the vertical axis and the outputs along the horizontal axis, with the transition probabilities within the same modality set to 1.
For transitions involving molecular internal information, which can be directly converted using specialized tools, we set the transition success rate to 1.
Molecular generation tasks are specifically addressed using SMILES representations, and the BLEU score serves as the metric for conversion effectiveness.
This standard is uniformly applied to all entries pertaining to internal information generation within the same row.
In cases where images can be converted from graphs using tools, the transition metrics for images are aligned with those for internal information.
For captioning and IUPAC naming tasks, the METEOR score is utilized as the conversion metric.
However, the molecular property prediction regression tasks can not be presented as probability.
We select classification tasks and use the mean ROC\_AUC as the conversion metric.
Finally, tasks involving text generation based on molecular properties are not within the scope of our research and are assigned a transition probability of zero.
This matrix provides a comprehensive view of modal adaptability across different task configurations, laying a foundational basis for future research.

\textbf{Comparing model architectures.}
Our research demonstrates the novel performance advantages of T5-based LLMs in molecular science, surpassing the capabilities of BERT and GPT variants.
We evaluate the performance of these models on molecular text-to-text and retrieval tasks.
We select the top five results for each task, providing a comparative analysis of the corresponding encoders and decoders.
As depicted in Figure \ref{fig:result}.b, we present the frequency of various encoders and the proportions of encoder-decoder combinations utilized.
T5-based models demonstrate a distinct advantage as encoders, while BERT models also show commendable performance.
Notably, the BioGPT \cite{luo2022biogpt}, a variant of the GPT model, excels in IUPAC name recognition tasks using captions and SELFIES, which is why it is prominently featured twice in our analysis.
In terms of decoding, T5-based models not only excel as encoder-decoder models, but they also perform robustly when deployed as decoders.
To assess the retrieval capabilities across different modalities, we consider all encoders and select the top five results for each task, as illustrated in Figure \ref{fig:result}.c.
We compare the embedding capabilities of different models and pooling strategies.
Additionally, we demonstrate the suitability of these modalities for cross-modal retrieval, using R@1 as the comparing metric.

\textbf{Modality and architecture insights.}
From the results, we present the following modality insights:
\begin{itemize}
    \item \textbf{IUPAC names for generation and captioning:}
    IUPAC names are conducive to text-based molecule generation and molecule captioning.
    \item \textbf{SMILES for IUPAC recognition:}
    SMILES representations are particularly well-suited for IUPAC recognition compared to other molecular internal information.
    \item \textbf{Preferred modalities for captioning:}
    Since some captions include the IUPAC name of the molecule, IUPAC is the most appropriate modality for molecule captioning, followed by SMILES.
    \item \textbf{Choosing modalities for retrieval:}
    For molecule retrieval tasks targeting captions, IUPAC is the optimal modality, followed by SMILES. Conversely, when the target is IUPAC names, SMILES becomes the most favorable modality.
    \item \textbf{Graph vs. textual modalities in property prediction:}
    In molecular property prediction tasks, among the selected top five rank combinations from 90 groups, graph modalities appear 40 times and SMILES 25 times, indicating a clear advantage for graph modalities and the relative superiority of SMILES over other textual representations.
\end{itemize}

As for the \textbf{model performance insights}:
\begin{itemize}
    \item \textbf{T5 series excellence:}
    The T5 series models show excellent performance across these tasks.
    \item \textbf{Graph models vs. T5 series:}
    Only the significantly larger T5 series models surpass GNNs, affirming the superiority of smaller-scale graph models for molecular embedding.
    \item \textbf{Efficacy of average pooling:}
    The average pooling mechanism tends to achieve superior performance more easily.
\end{itemize}

\subsection*{Case studies on model knowledge-learning preferences}

Our analysis aims to discern model preferences for knowledge acquisition.
To enhance interpretability, we specifically focus on the mapping processes from IUPAC names to captions and from SELFIES to captions, which bridge natural language tasks and chemical cross-domain tasks.
We analyze the text-inference processes of LLMs and compile two token mapping matrices.
To mitigate the impact of overwhelmingly frequent mappings on our analysis, we introduce a localized feature filtering method.
This approach allows us to analyze specific high-frequency token mappings that indicate knowledge-learning preferences.
By adjusting the threshold \(T\), we ensure that the selected token mappings are statistically significant.
Finally, we validate and analyze the effectiveness of our methods through molecular case studies, confirming their utility in real-world applications.

\begin{figure*}[t!]
\centering
\includegraphics[width=0.93\textwidth]{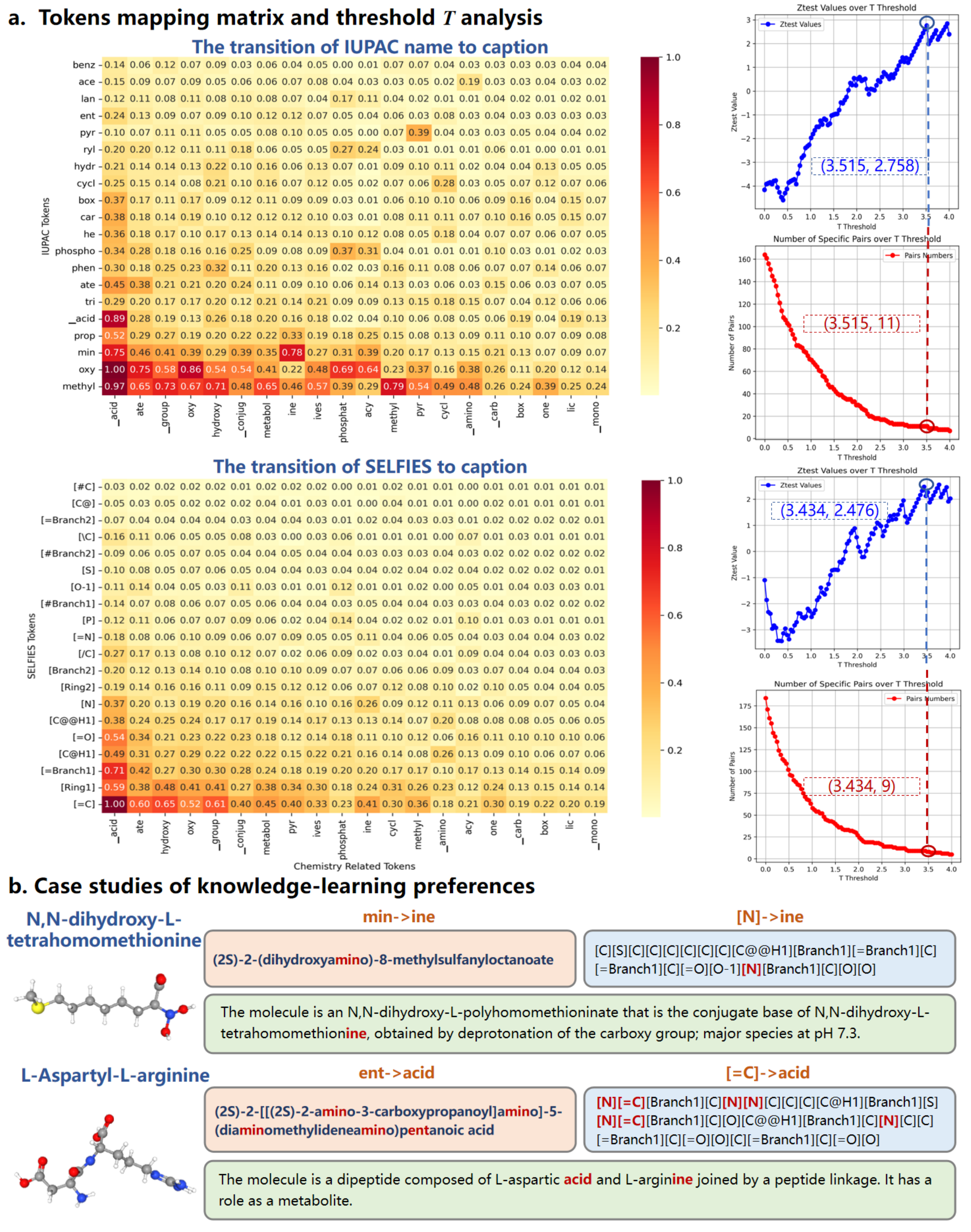}
\caption{
\textbf{Knowledge patterns and insights.}
\textbf{a. Tokens mapping matrix and threshold \(T\) analysis.} 
The two matrices represent the high-frequency tokens mapping patterns generated by the processes from IUPAC names and SELFIES to molecular captions. 
On the right of the figure, as the threshold \(T\) increases, the selection criteria for identifying specific high-frequency token pairs become more stringent, consequently reducing their number and impacting the significance levels.
\textbf{b. Case studies of knowledge-learning preferences.}
These cases are selected from model inference results, where the mapping of tokens exemplifies the model's preferences for knowledge learning.
}
\label{fig:Co-occurrence Matrix}
\end{figure*}

\textbf{Tokens mapping matrix.}
Initially, we extract typical tokens from the model's tokenizer, removing meaningless symbols and filtering the top 20 high-frequency chemical tokens to construct the mapping matrix \(A\).
Figure \ref{fig:Co-occurrence Matrix}.a displays the token mapping matrices after normalization and rearrangement based on the total counts of rows and columns.
The common high-frequency mappings, such as `oxy' and `methyl,' are prevalent across various tokens, indicating their widespread presence in chemical structures.
Traditional filtering methods, while capturing many of these frequent mappings, may inadvertently exclude tokens with specific contextual significance.
Our localized feature filtering method aims to refine this process by prioritizing tokens that provide greater contextual relevance, thereby enhancing the interpretation of our analysis.

\textbf{\added{Thresholds} \(T\) analysis.}
To identify particular high-frequency mapping pairs, we varied the threshold \(T\) to obtain different quantities of token mappings and their corresponding confidence levels.
We carefully balance the number of token mappings against the confidence levels to select an optimal \(T\), ensuring neither an overly small set of mappings nor insufficiently low confidence levels.
Taking the IUPAC name to caption process as an example, as shown in Figure \ref{fig:Co-occurrence Matrix}, at \(T = 3.515\), the Z-test attains a value of 2.758, corresponding to a confidence level of approximately 99.71\%.
This analysis identifies eleven specific token mapping pairs.
After excluding pairs with identical row and column names, we select seven unique pairs.
Similarly, for the SELFIES to caption process, at \(T = 3.434\), the Z-test reached a value of 2.476, translating to a confidence level of approximately 99.34\%.
From this, we identify five unique pairs.
By consolidating mappings with the same leading or trailing tokens, we categorize the mappings into the following 12 groups:
\begin{itemize}
    \item \textbf{IUPAC name to caption:}
    \begin{itemize}
        \item `benz': `group'
        \item `ace': `amino'
        \item `lan': `phosphat'
        \item `ent': `acid'
        \item `phen': `one'
        \item `acid': \{`box', `lic'\}
        \item `min': `ine'
    \end{itemize}
    \item \textbf{SELFIES to caption:}
    \begin{itemize}
        \item `[\textbackslash C]': \{`acy', `acid', `conjug'\}
        \item `[O-1]': \{`conjug', `phosphat'\}
        \item `[P]': \{`acy', `phosphat'\}
        \item `[N]': `ine'
        \item `[=C]': `acid'
    \end{itemize}
\end{itemize}
Our analysis elucidates the diversity and specificity of chemical token relationships, as demonstrated by the mapping patterns outlined.
These mappings not only facilitate a deeper understanding of chemical nomenclature across different representation methods but also reveal underlying scientific insights about molecular structure and function.

\textbf{Case studies of knowledge-learning preferences.}
Following these mapping patterns, we randomly select corresponding molecular IUPAC names, SELFIES, and their captions, as shown in Figure \ref{fig:Co-occurrence Matrix}.b. 
The token `min' in the IUPAC name is directly mapped to the `amino' group.
It reflects the NH2 group, which is common in biomolecules and crucial for biological functions.
The `[N]' in SELFIES denotes a nitrogen atom, key to the amino group’s bonding and interactions.
Similarly, the suffix `ine' in captions often signifies nitrogen-based compounds like amines and amino acids such as methionine, highlighting the amino group’s presence.
In another case study, the token `ent' often indicates pentane derivatives, such as `pentanoic' or `pentanoyl,' signaling a five-carbon chain integral to the molecule's structure.
For instance, in a dipeptide composed of L-aspartic acid and L-arginine, the presence of `pentanoic' denotes a five-carbon fatty acid essential for the molecule's stability and functional integrity. Concurrently, the SELFIES token `[=C]' underscores the significance of carbon double bonds, which are vital for maintaining molecular rigidity and facilitating biochemical interactions. Furthermore, the term `acid' in the caption specifically refers to the carboxylic acid groups in both amino acids, which are indispensable for forming peptide bonds that define the dipeptide's functionality.
\added{For more case studies, please refer to the Supplementary Figure 1.}

The model exhibits a robust understanding of molecular structure and functional groups, effectively mapping specific tokens to key chemical components such as carbon chains and double bonds.
It demonstrates the model's ability to discern critical elements that define molecular stability and reactivity, essential for predicting molecular behavior and interactions in biochemical contexts.
Through a series of model knowledge-learning insights, we demonstrate that the observed mapping patterns are not merely coincidental but reflective of the model's inherent ability to process and interpret complex chemical information.

\section*{Discussion}
\label{sec:Discussion}

\begin{figure*}[t!]
\centering
\includegraphics[width=\textwidth]{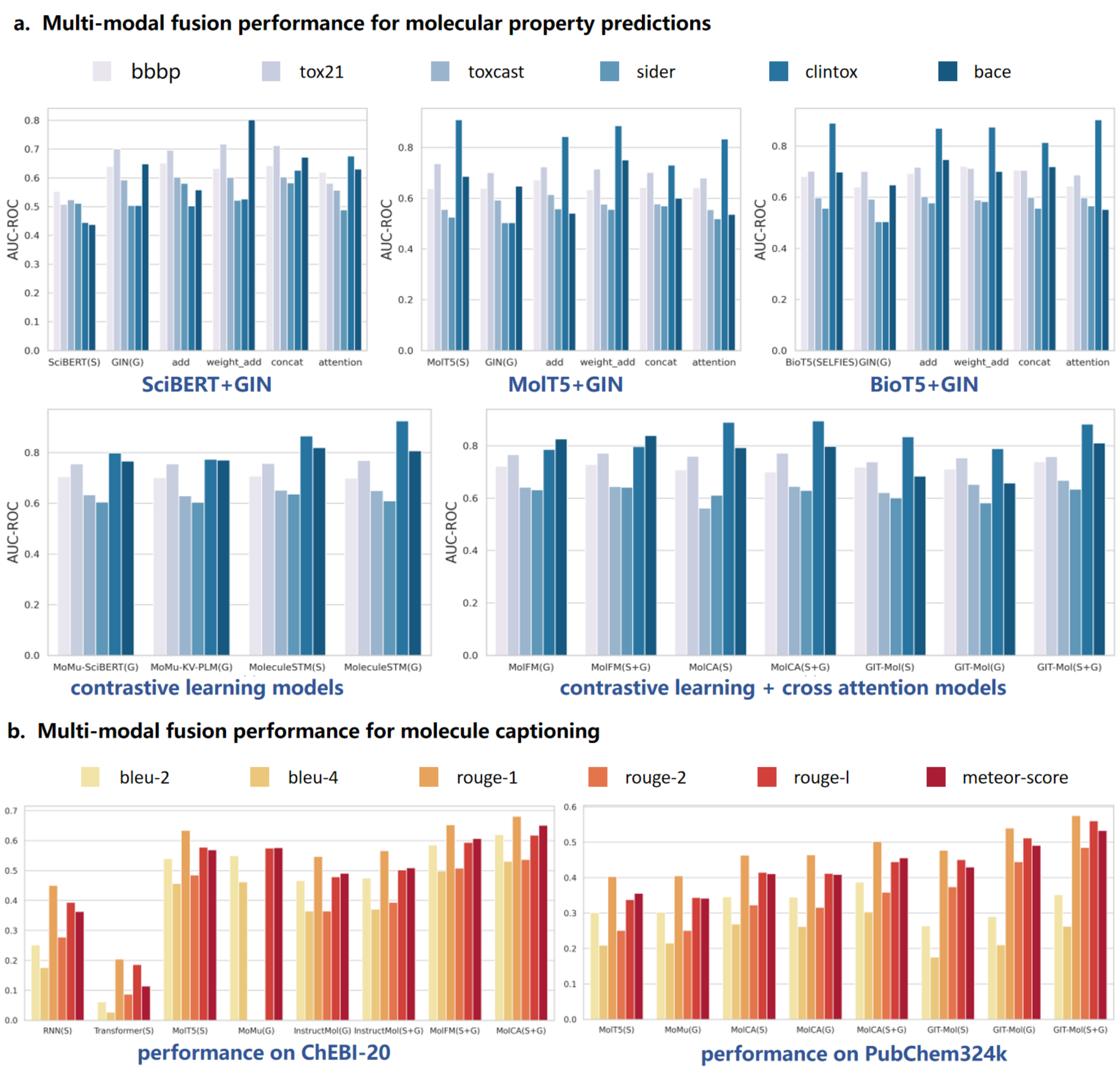}
\caption{\textbf{Performance of multi-modal fusion.}
\textbf{a. Multi-modal fusion performance for molecular property predictions.}
\added{This figure displays the AUC-ROC results for various molecular property prediction classification tasks.
It compares the performance of the SciBERT and MolT5 models as encoders using SMILES (S) as input text and BioT5 using SELFIES as input text, with the graph model (GIN) utilizing graph data (G).
In each subplot, the final results contributed by the vectors obtained after encoding and pooling from the foundation models are shown.
"add" represents vector addition, "weight\_add" represents adaptive weighted vector addition, "concat" represents concatenated encoding followed by pooling, and "attention" represents concatenated encoding processed by the attention mechanism before pooling.}
Different colors represent different tasks.
\textbf{b. Multi-modal fusion performance for molecule captioning.}
\added{This panel shows the performance of six textual similarity metrics across two common datasets for molecule captioning tasks.
The x-axis represents the models and input modalities, while the y-axis represents the metric values.}
Each color corresponds to a different metric.
}
\label{fig:multi_modal_result}
\end{figure*}

\textbf{Exploring the effects of multi-modal.}
To identify how data modalities influence model performance in molecular science tasks, we develop a multi-modal benchmark and carry out thousands of experiments.
Furthermore, we perform experiment replication and gather experimental data from other research to analyze the effects of factors related to multi-modal fusion.
\added{Figure \ref{fig:Conceptual-review} illustrates that molecular multi-modal fusion tasks include molecular retrieval, molecule captioning, and property prediction, with the latter two commonly used as benchmarks, as they integrate information from multiple modalities to improve molecular representation learning.
We use well-established benchmarks like MoleculeNet \cite{wu2018moleculenet} for property prediction, and ChEBI-20 \cite{edwards2022translation} and PubChem324k \cite{liu2023molca} for molecule captioning.
As shown in Figure \ref{fig:multi_modal_result}, our focus is on molecular property classification and molecule captioning, using SMILES, SELFIES, and graph representations processed by a graph isomorphism network (GIN) \cite{xu2018powerful}.
For more information about the results of multi-modal fusion, please refer to Supplementary Tables 10 and 11.}
The modal fusion techniques at the embedding layer include additive blending with weights, concatenation, and concatenation followed by self-attention.
In contrast, the encoding layer predominantly utilizes contrastive learning and cross-attention methods.
The comparative experiments based on pooling methods consistently utilize average pooling.

Our findings indicate that among the four fusion techniques at the embedding layer, only weighted additive blending consistently outperforms the baseline models, regardless of the underlying model and output modality.
Moreover, contrastive learning strategies also enhance model performance and are applicable to both graph and text models.
However, the use of cross-attention strategies tends to decrease performance in single-modality setups due to cross-modal interference while improving outcomes of modal fusion.
In the molecule captioning task, both contrastive learning and cross-attention mechanisms emerge as critical factors, with the choice of pretraining datasets also influencing outcomes.
Although our study predominantly focuses on prevalent data modalities, potentially overlooking some fingerprint modalities, our findings provide comparative insights and highlight areas for deeper investigation in future work.

\textbf{Scientific insights discovery.}
To explore the knowledge-learning preferences, we provide an interpretable analysis, revealing the foundational mechanisms in molecular SLM.
This analysis confirms the models' proficiency in encoding chemical knowledge, not only from the NLP perspective but also from a cross-domain viewpoint.
However, there remains substantial room for exploration.
In Figure \ref{fig:Co-occurrence Matrix}, we select knowledge patterns with a confidence level exceeding 99\%.
By lowering the threshold \(T\), although confidence may decrease, it provides an opportunity to uncover more novel insights, such as exploring unique properties of certain molecules or amino acids.
As models increase in complexity and the corpus of training data grows, the capacity of these systems to delve into and illuminate scientific phenomena also expands.
This enlargement not only enhances the depth and accuracy of the models' predictions but also allows for the exploration of previously unanticipated aspects of molecular behavior.
To more clearly showcase the models' capabilities in scientific understanding, future research should incorporate a broader spectrum of biochemical and physical chemistry knowledge.

\textbf{Evaluating molecular language modeling.}
A benchmark must exhibit a balanced distribution, preventing biases toward any specific molecular representation.
Analyzing the distribution of text lengths and tokenizer outputs helps confirm that a language model's tokenizer captures molecular information effectively.
Moreover, the diversity in scaffold distribution is crucial for determining the benchmark's generalization across chemical spaces, ensuring the language model can learn and predict a wide array of chemical entities.
Finally, the distribution and correlation of molecular descriptors reveal the complexity within the chemical space.
Overall, the visual analyses (Extended Data Fig. 1) of text length, scaffold count, and descriptor statistics provide insights into the dataset's characteristics, affirming the benchmark's robustness for tasks in molecular science.
While we strive to include a broad range of metrics for various tasks, our approach to molecular generation tasks primarily relies on similarity metrics in comparison to target texts.
\deleted{This methodology falls short in evaluating the actual volume of knowledge acquired by the models.
A more comprehensive assessment framework is needed to accurately gauge the learning capabilities of these models in molecular generation.}



In future research, we will focus on a sophisticated exploration of multi-modal data fusion to understand how various data modalities enhance model performance.
By incorporating scientific perspectives from biochemical and pharmacological fields, we expect to achieve a richer comprehension of the models' capabilities and potentially uncover more scientific insights.
We also plan to develop tailored evaluation metrics for various scientific language modeling tasks, moving beyond simple comparisons to target texts.
\deleted{These steps will bridge the gap between computational modeling and practical scientific applications, unlocking new potential in molecular sciences.}


\section*{Methods}
\label{sec:def}
In this section, we present essential background information.
We start by defining different molecular representation forms and vital LLMs, highlighting their significance in this field.
We then provide a detailed introduction to our benchmark dataset and the setup of our evaluation framework.
Finally, we elaborate on our statistically interpretable localized feature filtering method, which facilitates end-to-end analysis of model knowledge-learning preferences.

\subsection*{Molecular representation} 
Molecular structures can be represented in multiple ways for computational analyses, each with advantages and limitations.
These include graph structures, images, as well as one-dimensional notations. 
Figure \ref{fig:Conceptual-review}.a showcases examples of various data representation forms, while Figure \ref{fig:Conceptual-review}.b categorizes these representations into molecular internal and external information.
Additionally, it illustrates the relationships between these forms of representation and their corresponding tasks.

\textbf{Molecular internal information.}
The SMILES offers a widely-used one-dimensional notation for describing molecular structures.
InChI provides a unique identifier for chemical substances, facilitating the standard encoding of molecular information.
Further advancing molecular representations, SELFIES offers machine learning-oriented approaches, addressing the limitations of SMILES.
Moreover, graph structures naturally represent molecules, with atoms as nodes and bonds as edges, effectively capturing both local and global topological traits.

\textbf{Molecular external information.}
Images offer a visual 2D or 3D representation of molecules.
The IUPAC names provide a standardized naming system for organic compounds.
Additionally, molecular properties serve as a form of representation aiding human comprehension.
Furthermore, the molecular captions offer rich contextual information, encompassing textual descriptions of functions, properties, and inter-molecular relationships.

\subsection*{Large language models in molecular science}

\begin{figure*}[t!]
\centering
\includegraphics[width=\textwidth]{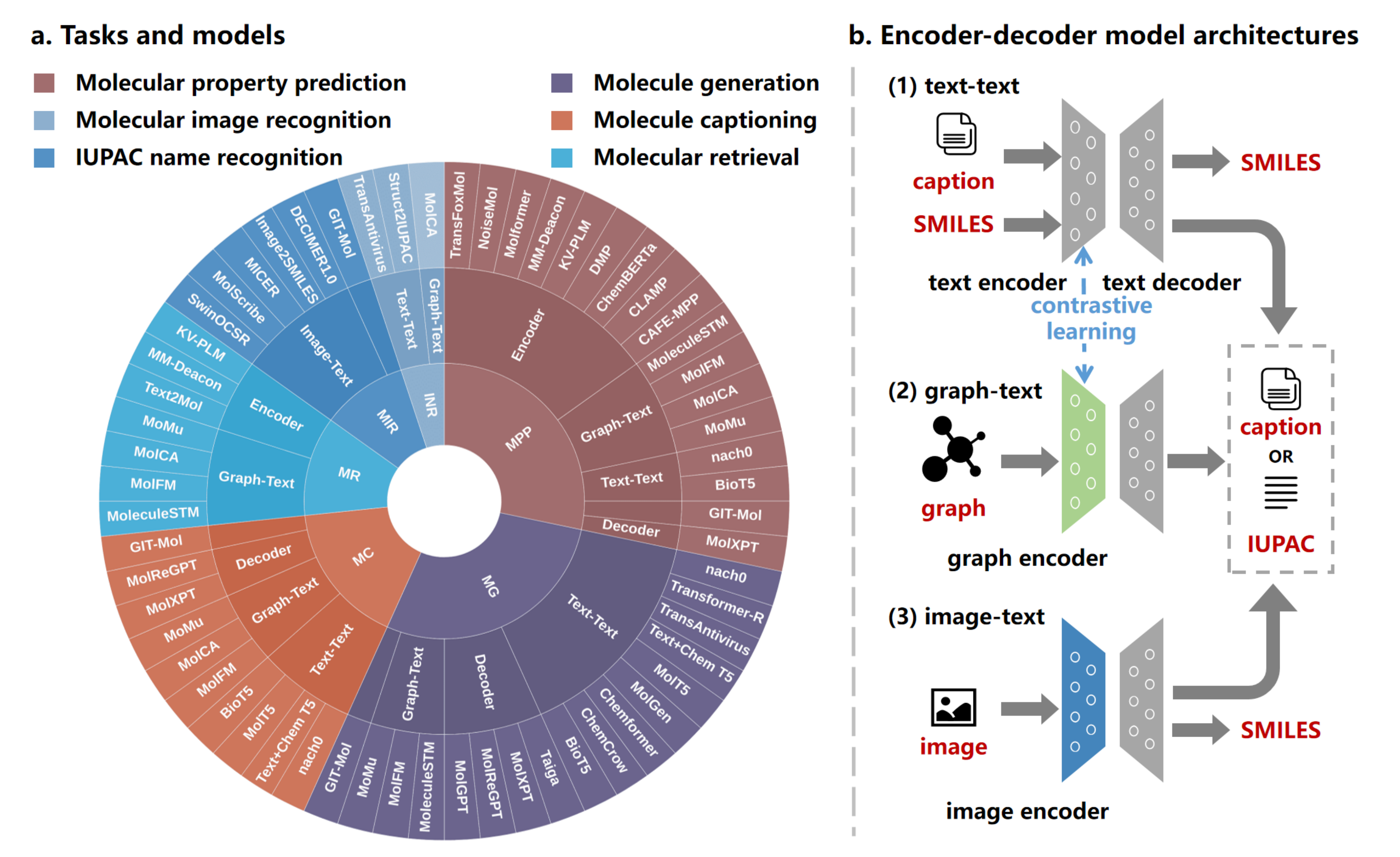}
\caption{\textbf{\added{An overview of model tasks and architectures.}}
    \textbf{\added{a. Tasks and models.}} 
    It clarifies the relationship between six downstream tasks and model architectures.
    \textbf{\added{b. Encoder-decoder model Architectures.}}
    It delineates three main frameworks: (1) \textbf{text-text} is primarily focused on text translation tasks;
    (2) \textbf{graph-text} is predominantly used in contrastive learning frameworks and serves as an encoder for downstream tasks;
    (3) \textbf{image-text} is chiefly applied in molecular image recognition tasks.
}
\label{fig:time-line}
\end{figure*}

Transformers, known for their self-attention mechanisms, enable the model to weigh the importance of different parts of the input data, which is crucial for understanding complex molecular structures of text modality.
\added{LLMs are Transformer-based models scaled to a large extent.
Among the notable foundation models in this domain are GPT \cite{radford2019language} and BERT \cite{kenton2019bert}.}
Especially in the fine-tuning stage, these models exhibit remarkable versatility and efficiency, requiring minimal task-specific adjustments for applications in molecular modeling.
\added{We present the classifications and architectures, as illustrated in Figure \ref{fig:time-line}.
The Details of the pivotal developments in transformer-based models for molecular modeling and design are shown in Supplementary Figure 2 and Table 2.}
To organize the models under a unified framework, we categorize them into three groups based on their architectures.
\begin{itemize}
    \item \textbf{Encoder-based models.}
    BERT and its derivatives like RoBERTa \cite{liu2019roberta} and domain-specific variants such as SciBERT \cite{beltagy2019scibert} and ChemBERTa \cite{chithrananda2020chemberta} have been effectively employed for molecular property prediction.
    The typical task can be mathematically represented as: $Encoder(X) = H$, where $X$ is the input data and $H$ denotes the encoded output.
    \item \textbf{Decoder-based models.}
    GPT-2 \cite{radford2019language} and ChatGPT \cite{schulman2022chatgpt} are superior in text generation capabilities. Specific-domain models like MolXPT \cite{liu2023molxpt}, based on GPT-2, and MolReGPT \cite{li2023empowering}, based on ChatGPT, showcase strong performance in generating molecular sequences and descriptions.
    The typical decoding task can be formulated as: $Y_t = Decoder(Y_{<t}, C)$, where \( Y_t \) is the output at time \( t \), \( Y_{<t} \) represents all preceding outputs, and \( C \) is the contextual input.
    \item \textbf{Encoder-decoder models.}
    The Transformer architecture itself exemplifies an end-to-end model, with T5 \cite{raffel2020exploring} and BART \cite{lewis2020bart} models further enhancing robust end-to-end text generation capabilities, advancing domain-specific applications.
    Beyond these text-based models, Encoder-decoder frameworks also extend to multi-modal models, accommodating a broader range of data inputs and analytical tasks.
    The encoder-decoder framework can be captured by the equation: $Y = Decoder(Encoder(X))$.
\end{itemize}


\textbf{Molecule captioning.}
It aims to develop textual descriptions based on given molecular representations.
It involves not only recognizing the molecular structure but also interpreting and conveying its functional properties, potential uses, and any relevant scientific context.
MolT5 \cite{edwards2022translation} is a groundbreaking model that uses the T5 architecture for end-to-end text tasks, achieving exceptional performance in single-modality models.
Following this, MolXPT \cite{liu2023molxpt}, built on GPT-2, utilizes a large corpus of PubMed titles and abstracts along with SMILES sequences from PubChem \cite{kim2023pubchem}.
It performs exceptionally well in few-shot learning and generative pre-training.
As for multi-modality models, MoMu \cite{su2022molecular} and MolFM \cite{luo2023molfm} employ cross-modal contrastive learning techniques.
MolCA \cite{liu2023molca} and GIT-Mol \cite{liu2023git} combine Q-Former \cite{li2023blip} with contrastive learning to integrate SMILES strings and graphs, leveraging LLMs' generative capabilities for captioning.

\textbf{IUPAC name recognition.}
Unlike free-form textual descriptions, IUPAC names possess the advantage of uniformity and specificity, making them highly structured and predictable.
Furthermore, the task of IUPAC name recognition involves not just the decoding of these structured names but also understanding the underlying molecular structure they represent.
Struct2IUPAC \cite{krasnov2021struct2iupac} employs a Transformer-based architecture, which is complemented by a specialized tokenizer adept at processing both SMILES strings and IUPAC nomenclature.
Moreover, the TransAntivirus \cite{mao2023transformer} goes a step further by adopting the versatile T5 model architecture.

\textbf{Molecular property prediction.}
Molecular representation is critical in molecular modeling, providing ways to describe and encode molecular structures for computational use.
Within this domain, molecular property prediction is essential, assessing attributes like solubility, toxicity, and protein binding affinity.
Based on BERT, KV-PLM \cite{zeng2022deep} employs a dual tokenizer to better recognize specific atoms and functional groups within SMILES strings.
As for multi-modal methods, MoMu \cite{su2022molecular} applies contrastive learning techniques to assimilate modality-specific and complementary information from text and graph data. 
Post-encoding, embeddings are transformed into fixed-length vectors via a pooling layer, employing methods like average pooling and max pooling.
\textbf{Average pooling} calculates the mean of all sequence vectors for a general context representation. 
\textbf{Max pooling} selects each feature's maximum value across the sequence, highlighting key molecular features.

\textbf{Molecular retrieval.}
Molecular retrieval aims for efficient, accurate identification of molecules in large datasets.
Its effectiveness relies on the type of molecular data used, like structural formulas or SMILES strings.
In cross-modal retrieval tasks, Text2Mol \cite{edwards2021text2mol} excels by combining natural language with molecular data into a unified semantic space for effective retrieval.
Similarly, MoMu \cite{su2022molecular} and MolFM \cite{luo2023molfm} use advanced contrastive learning to merge text and molecular graphics, enhancing the feature space for robust retrieval tasks.
As for the pooling layers, similar to those detailed in molecular property prediction.

\textbf{Molecular image recognition.}
In our research, molecular image recognition focuses on converting visual representations of molecular structures into standardized, textual formats like SMILES.
This task is particularly crucial for automating the interpretation of chemical literature, where molecular images are frequently used to describe compounds.
DECIMER 1.0 \cite{rajan2021decimer}, Image2SMILES \cite{khokhlov2022image2smiles} and MICER \cite{yi2022micer} employ Transformer-based architectures and image augmentation, handling large datasets effectively.
SwinOCSR \cite{xu2022swinocsr} utilizes Swin Transformers \cite{liu2021swin} for global feature extraction and tackles token imbalance with focal loss.

\textbf{Molecule generation.}
Our research focuses on developing one-dimensional textual representations, particularly in SMILES format.
This includes text-based de novo molecule generation and molecular optimization due to their methodological similarities.
In molecular generation, models adept at molecule captioning are equally capable of molecule generation.
BioT5 \cite{pei2023biot5} further expands this to include protein-related texts, focusing on SELFIES and extending to protein property and drug-target interaction prediction.
Molecular optimization aims to refine molecules for improved properties, aligning with generative goals.
MoleculeSTM \cite{liu2023multi} employs contrastive learning on graphs and text, optimizing target attributes like binding affinity and drug-relatedness with precision.

\subsection*{ChEBI-20-MM and experimental setup}
\begin{figure*}[t!]
\centering
\includegraphics[width=\textwidth]{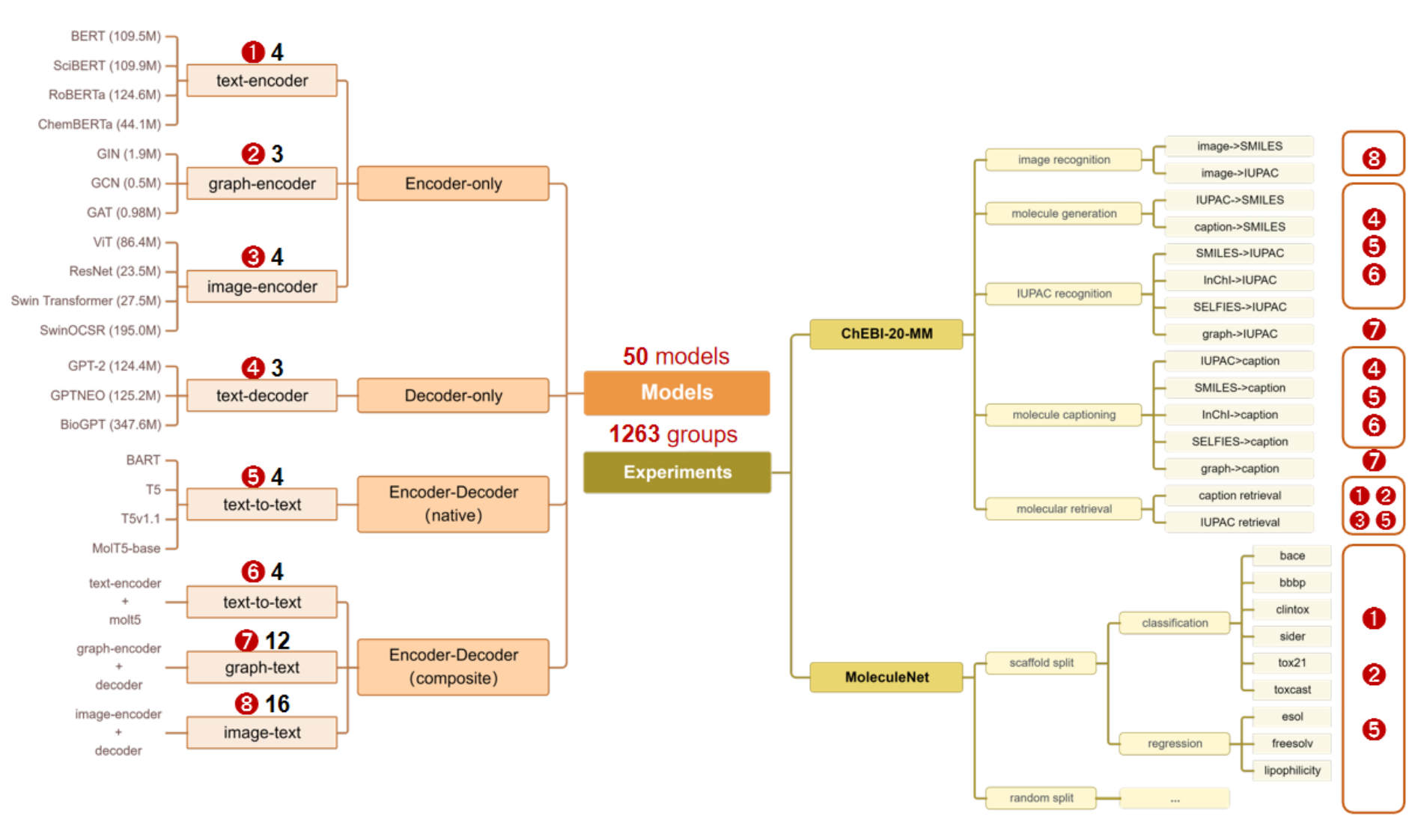}
\caption{\textbf{Benchmark experiments overview.}
\added{Our study encompasses tests across eight primary model architectures, each featuring 3-4 common foundation models or composite models within its category.}
In total, \textbf{1263} experimental setups were conducted, demonstrating the adaptability of various model architectures to different task types.
\added{The figure annotates the parameter size of foundation models and labels with items (1-8) to indicate architecture patterns suited for specific tasks.}}
\label{fig:benchmark}
\end{figure*}

As interest in AI for molecular research continues to grow, an increasing array of datasets is coming into focus.
We provide a detailed overview of common molecular datasets in Supplementary Table 1.
Meanwhile, molecular representations, particularly molecular text descriptions which are typically scarce, are now seeing increased utilization with MolT5's ChEBI-20 dataset in applications such as molecule generation and captioning.
To facilitate multi-modal data evaluation, we expand our data modality variety, introducing ChEBI-20-MM.
This benchmark effectively evaluates the versatility of data types and model combinations in describing, generating, and embedding molecular data.
\added{Figure \ref{fig:benchmark} showcases our evaluation of prevalent foundation models and experiments.}
For comparing models' embedding capabilities, our experiments include datasets from MoleculeNet, enabling thorough assessments in biochemical and physicochemical contexts.

\textbf{Dataset construction.}
The ChEBI-20 dataset includes molecular SMILES and captions.
Building on this, we use tools such as RDKit to supplement the dataset with InChI, graphs, SELFIES, and other internal information based on the SMILES.
Additionally, IUPAC names and images are sourced from PubChem.
During the dataset curation process, we maintain the original ChEBI-20 split of 8:1:1 (train: validation: test) and remove molecules that can not be converted to SELFIES to ensure the validity of all molecular data.
In total, the refined dataset comprises 32,998 molecules, which are suitable for the text generation tasks depicted in Figure \ref{fig:Conceptual-review}, including text generation based on multi-modal fusion.

\textbf{Data distribution.}
We analyze the \textbf{suitability} of data sources for language models and \textbf{chemical space coverage}.
As shown in Extended Data Fig. 1, we use different visualization methods to analyze the distribution of text lengths and the number of tokens generated by each model's tokenizer for various text data types.
This approach allows us to evaluate the adaptability of language models to the textual characteristics of our dataset.
We also focus on the top 10 scaffolds within the dataset, counting the number of molecules for each scaffold.
Here, semi-transparent bars represent the total count, while solid bars indicate the quantity in the training set.
On the other hand, for the analysis of \textbf{chemical space coverage}, we choose molecular weight (MW), LogP, the number of aromatic rings, and the Topological Polar Surface Area (TPSA) as descriptors.
We examine the distribution and correlation of these descriptors within the dataset, providing insights into the chemical diversity and complexity present in our data.

\textbf{Experimental setup.}
We conduct 1263 experiments, encompassing 291 generative and retrieval tasks from the ChEBI-20-MM benchmark and 972 property prediction experiments.
For nine text-to-text tasks, our models include 11 architectural variations (items 3, 4, 5 in Figure \ref{fig:benchmark}), combining three decoder-only models, four encoder-decoder models, and four composite models with MolT5 as the decoder paired with various encoders.
We conduct experiments based on the setup, performing each experiment three times to calculate the mean and standard deviation of the results. 
For nine text-to-text translation tasks, our models encompass 11 architectural variations (as illustrated in Figure \ref{fig:benchmark}, items 4, 5, 6), which include three decoder-only models, four encoder-decoder models, and four composite models featuring MolT5 as the decoder paired with various encoders.
For two graph-to-text tasks, we deploy 12 distinct architectures (item 7), and for two image-to-text tasks, we utilize 16 different architectures (item 8).
Additionally, for two retrieval and nine property prediction tasks, we implement 11 encoder architectures (items 1, 2, 3, 5) alongside native encoder-decoder configurations serving as molecular encoders.

\textbf{Metrics.}
We use a combination of BLEU-2, BLEU-4, ROUGE-1, ROUGE-2, ROUGE-L, and METEOR scores for IUPAC name recognition and molecule captioning.
Moreover, for image recognition and molecule generation, we assess models using BLEU for textual similarity, Exact Match for precision, Valid for chemical validity of generated molecules, Levenshtein distance for structural similarity, and fingerprint-based metrics such as MACCS, RDKit, and Morgan for capturing chemical feature representation.
Molecular retrieval is evaluated using mean reciprocal rank (MRR) and Recall at K (R@1, R@5, R@10).
For the property prediction metrics, We utilize widely recognized classification and regression tasks from MoleculeNet. 
For classification, we measure the ROC\_AUC, the area under the precision-recall curve (PR\_AUC), and the mean F1 score (F1\_score).
For regression tasks, we use mean squared error (MSE\_mean), root mean squared error (RMSE\_mean), and mean absolute error (MAE\_mean) to assess the predictive accuracy of the models quantitatively.

\subsection*{Analysis of model knowledge-learning preferences}
The interpretability methods have become more important in LLMs, particularly in molecular science.
As research increasingly validates the effectiveness of these LLMs, a significant challenge remains their `black box' nature.
This opacity limits our understanding of what knowledge the models have actually learned.
Furthermore, it restricts the design and application of these models.
While existing interpretability methods, such as case study-based attention and embedding visualizations, offer some insights, they often lack generalizability and provide limited understanding to researchers.
Therefore, there is a pressing need for a unified analysis of batch data that can extract and distill insights into the models' knowledge-learning preferences and present them in an interpretable textual format.
To explore the knowledge-learning preferences of models in the process of molecular SLM, this section introduces a language-driven,  statistically significant, interpretable method.
This tool constructs a token mapping matrix between input data and inference outcomes, revealing the model's mechanism.

\textbf{Particular high-frequency token mappings.}
We decompose the input data and inference outcomes of LLMs into corresponding tokens and select high-frequency tokens to construct a mapping matrix, denoted as \(A\).
Additionally, we apply special processing to these tokens, such as filtering out tokens related to chemical vocabulary in captions and removing irrelevant words.
We typically view high-frequency token mappings as indicative of specialized knowledge.
Our approach identifies two types of high-frequency token mappings: \textbf{general pairs}, which are widespread and lack specific context, and \textbf{particular pairs}, essential for certain contexts but not always the most frequent.
The challenge arises when standard analysis methods prioritize these general high-frequency mappings, consequently overlooking the specific ones that often carry more valuable contextual insights.

\textbf{Localized feature filtering method.}
We utilize a localized feature filtering method for high-frequency selection to eliminate common high-frequency mapping pairs, thus more accurately filtering the model's knowledge mappings.
\added{Subsequently, a statistical significance method is applied to determine the filtering threshold, identifying high-frequency mapping pairs that reflect the model's chemical knowledge.}

\textbf{Proof Point 1:} Sorting the matrix \(A\) in descending order by row and column totals enhances the similarity among adjacent values.
Assuming each \(A_{ij}\) follows an independent and identical distribution, such as normal distribution, the row sums \(R_i\) and column sums \(C_j\) mimic this distribution.
Consequently, in the sorted matrix \(A'\), elements next to each other are more likely to come from rows or columns with similar totals.

Considering \(A'_{kl}\) and its neighbors \(A'_{k\pm1,l\pm1}\) in \(A'\), which were close in the sum in the original matrix \(A\), the probability they are similar is mathematically represented as:

\begin{equation}
P(A'_{kl} \approx A'_{k\pm1,l\pm1} | R_i \approx R_{i\pm1}, C_j \approx C_{j\pm1}) > P(A_{ij} \approx A_{i\pm1,j\pm1})
\label{eq:probability_condition}
\end{equation}

This highlights that adjacent elements in \(A'\) are more likely to have close values compared to those in \(A\), due to the sorting based on similar row and column sums.

\textbf{Proof Point 2:} To uncover significant patterns within a matrix, we introduce a threshold \(T\), aiming to spotlight elements \(A_{ij}\) that significantly surpass the mean plus a certain number of standard deviations of their neighboring values.
This process starts with assuming elements in matrix \(A\) are drawn from a distribution, enabling the calculation of each element's neighborhood mean \(\mu_{neighbor}\) and standard deviation \(\sigma_{neighbor}\).
This step evaluates whether an element exceeds the set threshold, the condition to filter elements that starkly differ from their surrounding context:

\begin{equation}
A_{ij} > \mu_{neighbor} + T \cdot \sigma_{neighbor}
\label{eq:threshold_condition}
\end{equation}

To assess the actual significance of these findings, we calculate the actual proportion \(P_{actual}\) of elements exceeding this threshold across the matrix and compare it to the expected proportion \(P_{expected}\) under a normal distribution.
The actual and expected proportion is determined by the fraction of elements satisfying our condition:

\begin{equation}
P_{actual} = \frac{\sum I_{ij}}{n \times m}
\label{eq:actual_probability}
\end{equation}

\begin{equation}
P_{expected} = 1 - \Phi\left(\frac{\mu_{neighbor} + T \cdot \sigma_{neighbor} - \mu}{\sigma}\right)
\label{eq:expected_probability}
\end{equation}

where \(I_{ij}\) is an indicator function that equals 1 when the condition is met and 0 otherwise, and \(n \times m\) represents the matrix size. 
The \(\Phi\) represents the cumulative distribution function of the standard normal distribution, and \(\mu\) and \(\sigma\) are the mean and standard deviation of the entire matrix's elements.

Finally, a statistical test, such as the Z-test, evaluates the significance of the observed proportion against the expected, indicating whether the identified patterns are indeed statistically significant beyond random chance:

\begin{equation}
Z = \frac{P_{actual} - P_{expected}}{\sqrt{\frac{P_{expected}(1 - P_{expected})}{n \times m}}}
\label{eq:standard_score}
\end{equation}

This succinct methodology not only clarifies the process of identifying statistically significant patterns but also underlines its effectiveness in distinguishing genuine high-frequency elements from those of a random distribution, offering valuable insights into the structure of the knowledge-learning preferences of models.

\textbf{Comparison with other interpretability methods.}
Interpretability in machine learning, especially within molecular science, utilizes various methods to render the decision-making processes of models both accessible and comprehensible.
\added{As shown in Supplementary Table 12, researchers frequently employ techniques such as text-based attention map visualizations and molecular graph-based attention heatmaps.}
These methods typically rely on inductive reasoning from specific case studies, employing a bottom-up approach to derive insights.
While these methods can provide valuable insights, their dependence on extensive case studies renders the process time-consuming and complex.
Furthermore, the insights obtained through these traditional methods often require further verification to confirm their validity, thus challenging their reliability and general applicability.
\added{In contrast, our approach, termed the localized feature filtering method, initiates the process of insight derivation through statistical significance techniques.}
We can set thresholds to balance the volume of insights with their confidence levels, employing a top-down validation approach to substantiate the effectiveness of our selected insights.
This method not only ensures the credibility of the insights but also significantly boosts the efficiency of insight extraction.
By melding rigorous statistical analysis with a systematic validation framework, our approach enhances the interpretability of molecular models, offering a robust and efficient alternative to traditional, more heuristic methods.

\textbf{Utility and potential application.}
The LLMs demonstrate significant superiority in molecular science.
While their performance in representation tasks may align with that of traditional approaches and graph models, their unparalleled ability to generate coherent and contextually relevant text sets them apart.
Our benchmark and modal transition matrix are designed to rapidly identify optimal molecular modalities, model architectures, and training strategies for specialized tasks within molecular science.
For instance, the BioT5+ model \cite{pei2024biot5+}, which utilizes molecular IUPAC names and SELFIES for T5-based caption generation, achieves state-of-the-art results.
It illustrates a new use of IUPAC names, a previously unrecognized strategy in the field, which is also concluded by our benchmark analysis.

The localized feature filtering method facilitates the rapid extraction of a model's knowledge-learning preferences.
This capability is instrumental in guiding model training, for example, by optimizing the tokenizer or refining model parameters to boost learning efficiency.
Furthermore, it allows for the strategic adjustment of the threshold \(T\) to unearth potential scientific insights.
By exploring these insights, researchers can probe new dimensions of molecular behavior, paving the way for innovative breakthroughs in molecular design and synthesis.
Through these applications and methodological advancements, our approach demonstrates its utility, providing a robust framework for enhancing both the interpretability and functionality of models in molecular science.

\section*{Acknowledgments}
This work is supported by grants from the National Natural Science Foundation of China (61902446, 62172456, and 91937302), the Major Key Project of PCL PCL2021A13, and Peng Cheng Cloud-Brain.

\section*{Author Contributions Statement}
Pengfei Liu: Conceptualization, Methodology, Data curation, Model training, Writing-original draft.
Jun Tao: Funding acquisition, Writing review editing.
Zhixiang Ren: Conceptualization, Formal analysis, Supervision, Funding acquisition, Writing review editing.

\section*{Competing Interests Statement}
The authors declare that they have no known competing financial interests or personal relationships that could have appeared to influence the work reported in this paper.

\section*{Data availability}
The dataset ChEBI-20-MM \cite{pengfei_liu_2024} can be accessed at \url{https://huggingface.co/datasets/liupf/ChEBI-20-MM}.

\section*{Code availability}
The open-source project repository \cite{pengfei_liu_2024_14293309} is available in GitHub at \url{https://github.com/AI-HPC-Research-Team/SLM4Mol}.
It is available for non-commercial use.

\bibliography{arxiv.bib}






\clearpage

\captionsetup[figure]{labelformat=empty}
\begin{figure*}[t!]
\centering
\includegraphics[width=\textwidth]{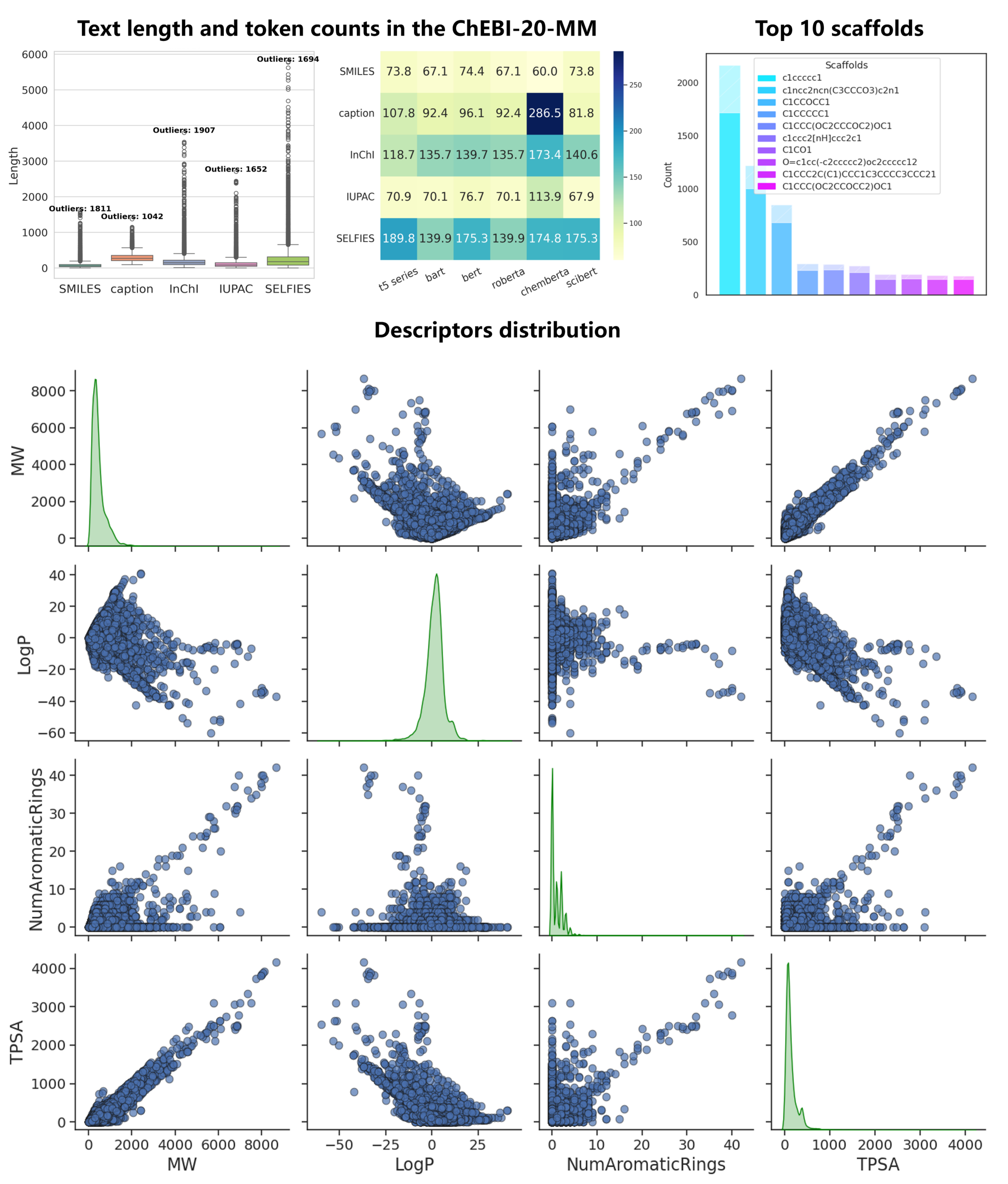}
\caption*{Extended Data Fig. 1| Visualization of data source suitability and chemical space diversity.}
\label{fig:org_data}
\end{figure*}


\clearpage

\tableofcontents








\clearpage

\section{The databases and datasets for molecular science}
In this section, we highlight vital data sources essential for molecular science research.
As detailed in Supplementary Table 1, PubChem and ZINC databases provide extensive compounds for virtual screening.
ChEMBL offers comprehensive data on bioactive molecules, while ChEBI focuses on cataloging small chemical compounds.
DrugBank serves as a valuable resource for drug and target information.
Lastly, the USPTO database, encompassing patent filings and research data, forms a vital part of the scientific literature.
\begin{table}[ht]
\centering
\caption*{\textbf{Supplementary Table 1:} Overview of data sources for molecular science}
\label{tab:databases}
\resizebox{\textwidth}{!}{%
\begin{tabular}{l|l|lll}
\toprule
\textbf{Type} & \textbf{Data Source}  & \textbf{Content} & \textbf{No. of Entries} & \textbf{Description} \\
\midrule
\multirow{6}{*}{Database} 
&\href{https://pubchem.ncbi.nlm.nih.gov/}{PubChem}  \cite{kim2023pubchem}  & compounds and substances & 116M & molecule database \\
&\href{https://zinc20.docking.org/}{ZINC}  \cite{irwin2020zinc20}  & compounds and substances & 230M & virtual screening database \\
&\href{https://www.ebi.ac.uk/chembl/}{ChEMBL}  \cite{zdrazil2023chembl}  & compounds with text descriptions & 2.4M & bioactive molecule database \\
&\href{https://www.ebi.ac.uk/chebi/aboutChebiForward.do}{ChEBI}  \cite{hastings2016chebi}  & small chemical compounds & 61,032 & molecular entity dictionary \\
&\href{https://go.drugbank.com/}{DrugBank}  \cite{wishart2018drugbank}  & small molecule drugs & 12,695 & drug/target database \\
&\href{https://www.uspto.gov/}{USPTO} & patent filings and data & —— & scientific corpus \\
\hline
\multirow{13}{*}{Dataset} 
&\href{https://www.tensorflow.org/datasets/catalog/c4}{C4}  \cite{raffel2020exploring}  & Clean English text & —— & Pretrain \\
&MoMu dataset  \cite{su2022molecular}  & Graph-document pairs & 15,613 & Pretrain \\
&\href{https://paperswithcode.com/dataset/moses}{MOSES}  \cite{polykovskiy2020molecular}  & Molecular structures & 1.9M & Generation \\
&\href{https://github.com/thunlp/KV-PLM}{PCdes}  \cite{zeng2022deep}  & SMILES-text descriptions & 1.5K & Molecular retrieval \\
&\href{https://github.com/cnedwards/text2mol/tree/master/data}{ChEBI-20}  \cite{edwards2021text2mol} & SMILES-text descriptions & 33,010 & Generation and captioning \\
&\href{https://huggingface.co/datasets/chao1224/MoleculeSTM/tree/main/PubChemSTM_data}{PubChemSTM}  \cite{liu2023multi}  & SMILES-text descriptions & 281K & Generation \\
&\href{https://github.com/zjunlp/Mol-Instructions}{Mol-Instructions}  \cite{fang2023mol}  & SMILES-text descriptions & 333K & Generation and captioning \\
&\href{https://github.com/deepchem/moleculenet}{MoleculeNet}  \cite{wu2018moleculenet} & Molecular properties. & 700K & Property prediction \\
&\href{https://sourceforge.net/p/osra/wiki/Validation/}{USPTO(OSRA) \cite{osra-validation-2020}} & Images and ground truth & 5,719 & Image recognition \\
&\href{http://www.ifs.tuwien.ac.at/~clef-ip/download/2012/index.shtml}{CLEF \cite{clef-ip-2012}}  & Images and ground truth & 865 & Image recognition \\
&\href{https://sourceforge.net/p/osra/wiki/Validation/}{JPO \cite{osra-validation-2020}}  & Images and ground truth & 450 & Image recognition \\
&\href{https://www.cs.bham.ac.uk/research/ groupings/reasoning/sdag/chemical.php}{UOB \cite{molrecuob-2020}}  & Images and ground truth & 5,740 & Image recognition \\
&\href{https://paperswithcode.com/dataset/uspto-50k}{USPTO-50k}  \cite{schneider2016s} & Chemical reactions & 50K & Reactions prediction \\
\bottomrule
\end{tabular}%
}
\end{table}

Moreover, various vital datasets support a wide range of research, from pretraining to specific molecular science tasks.
As shown in Supplementary Table 1, the Colossal Clean Crawled Corpus (C4) dataset offers a vast amount of clean English text scraped from the web for the pretraining of the T5 model.
MOSES provides molecular structures from ZINC, while MoleculeNet encompasses a broad spectrum of molecular property predictions ranging from quantum mechanics to biophysics and physiology.
Additionally, datasets such as USPTO, CLEF, JPO, and UOB offer molecular images with ground truth, supporting image recognition tasks.
The PCdes, ChEBI-20, and PubChemSTM datasets, along with Mol-Instructions, provide essential SMILES-text pairs that facilitate cross-information retrieval, molecule generation, and molecular captioning tasks.

\section{Supplementary case studies on the knowledge-learning preferences of models}

This section presents additional application cases for our localized feature filtering method beyond the SELFIES to caption and IUPAC to caption tasks discussed in the manuscript.
The method can support a variety of molecular modalities, all of which are text-generation tasks that can benefit from this analysis.
As shown in Figure 1 of the manuscript, our method is applicable to tasks such as molecule captioning, IUPAC name recognition, and molecule generation.
Due to the relatively low frequency of InChI usage, we present only a few modalities, such as SMILES, SELFIES, IUPAC names, and captions.
As illustrated in Figure 3 of the manuscript, we have already demonstrated the SELFIES to caption and IUPAC to caption tasks.
Furthermore, as illustrated in Supplementary Figure 2, we analyze the recognition of IUPAC names and the generation of molecules. We also present insights into the model's knowledge-learning preferences, which include Z-test results, confidence levels, and specific high-frequency token mappings.

\begin{figure*}[t!]
\centering
\includegraphics[width=\textwidth]{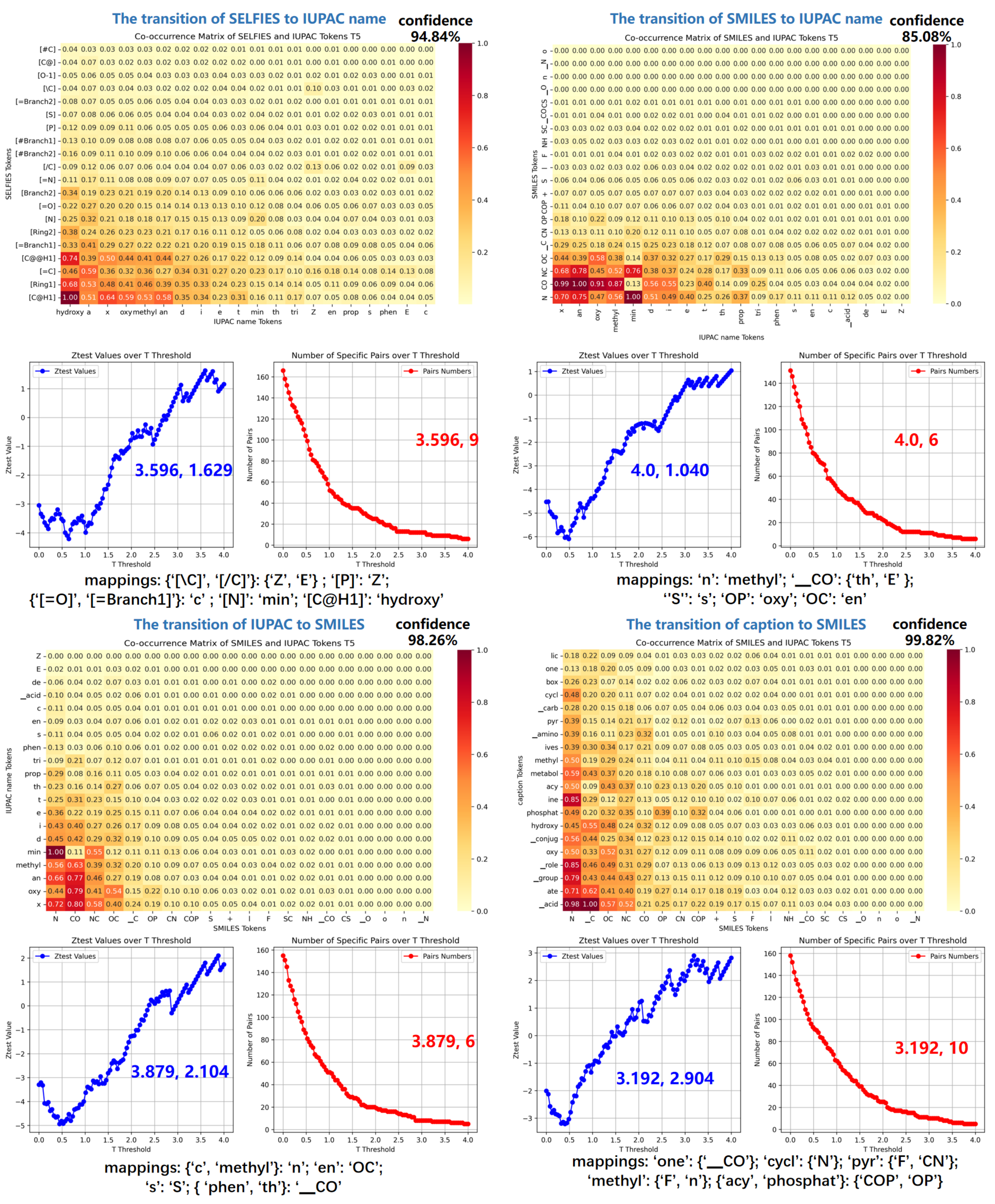}
\caption*{\textbf{Supplementary Figure 1: }
    Tokens mapping matrix and threshold T analysis of IUPAC name recognition and molecule generation.
    Each subplot for the tasks displays a tokens co-occurrence matrix, illustrating the degree of mapping between key tokens across different modalities.
    The analysis is complemented by the T threshold derived from the localized feature filtering method, along with corresponding Z-test results, confidence, and insights into particular high-frequency token mappings.
}
\label{fig:supplementary_information_figure_1}
\end{figure*}

\clearpage

\section{The details of models}
We present a timeline showcasing the pivotal developments in transformer-based models for molecular modeling and design, alongside classifications and architectures, as illustrated in Supplementary Figure 1.
An overview of model classifications and architectures is shown in Supplementary Table 2.
\begin{figure*}[!h]
\centering
\includegraphics[width=\textwidth]{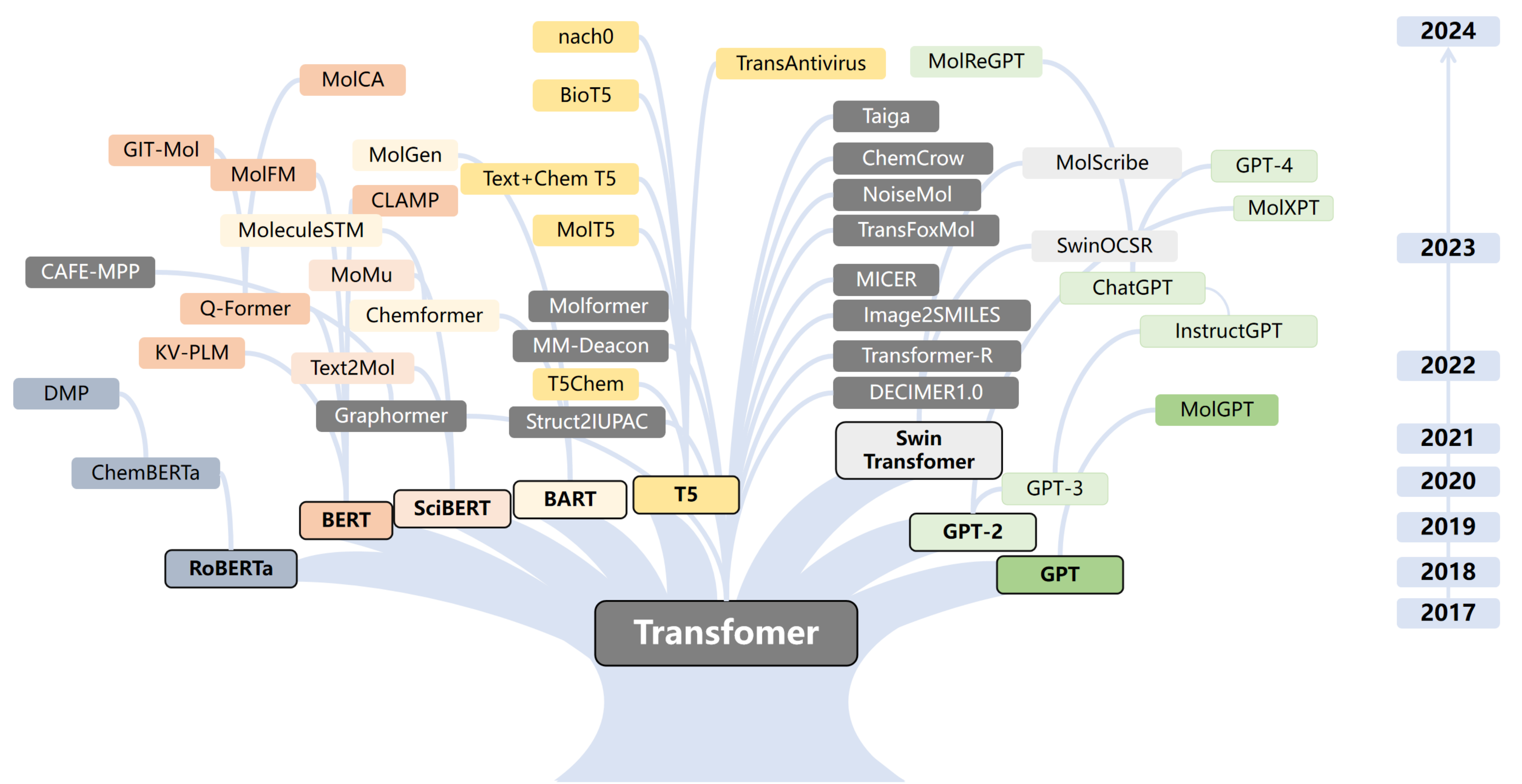}
\caption*{\textbf{Supplementary Figure 2: }
    Timeline of the key developments of LLMs.
    The models based on the Transformer architecture are differentiated by various colors to denote that they are based on the nine foundation models.
}
\label{fig:time-line_1}
\end{figure*}

\begin{table}[ht]
  \centering
  \caption*{\textbf{Supplementary Table 2:} Overview of models in molecular modeling design categorized by their architecture and backbone.
  The abbreviations \textbf{MC}, \textbf{INR}, \textbf{MPP}, \textbf{MR}, \textbf{MIR}, \textbf{MG}, and \textbf{OT} correspond to the tasks investigated: molecule captioning, IUPAC name recognition, molecular property prediction, molecular retrieval, molecular image recognition, molecule generation, and other tasks, respectively.
  Each {\Large $\bullet$} symbol in the table denotes the demonstrated task capabilities of the models as reported in their publications.
  Notably, the {\Large \(\circ\)} symbol specifically signifies the specialized task of molecular optimization within the broader context of molecule generation.}
  \label{tab:models_review}
  \resizebox{\textwidth}{!}{%
  \begin{tabular}{|c|c|c|c|c|c|c|c|c|c|c|}
    \hline
    \textbf{Architecture} & \textbf{Backbone} & \textbf{Model} & \textbf{Date} &  \textbf{MC} & \textbf{INR} & \textbf{MPP} & \textbf{MR} & \textbf{MIR} & \textbf{MG} & \textbf{OT} \\
    \hline
    \multirow{10}{*}{Encoder-only} & \multirow{2}{*}{BERT} & \href{https://github.com/thunlp/KV-PLM}{KV-PLM}  \cite{zeng2022deep} & 2022/02 & & & {\Large $\bullet$} & {\Large $\bullet$} & & & {\Large $\bullet$} \\
    & & \href{https://github.com/ml-jku/clamp}{CLAMP} \cite{seidl2023enhancing} & 2023/05 &   &  & {\Large $\bullet$} &  &  &  & \\
    \cline{2-4}
    & \multirow{2}{*}{RoBERTa} & \href{https://huggingface.co/seyonec/ChemBERTa-zinc-base-v1}{ChemBERTa} \cite{chithrananda2020chemberta} & 2020/10 &  &  & {\Large $\bullet$} &  &  &  & \\
    & & DMP \cite{zhu2023dual} & 2021/10 &  &  & {\Large $\bullet$} &  &  &  & \\
    \cline{2-4}
    & \multirow{4}{*}{Transformer} & \href{https://github.com/IBM/molformer}{Molformer} \cite{ross2022large} & 2022/04 &  &  & {\Large $\bullet$} &  &  &  & \\
    & & MM-Deacon \cite{guo2022multilingual} & 2022/04 &  &  & {\Large $\bullet$} & {\Large $\bullet$} &  &  & \\
    & & \href{https://github.com/Jiangjing0122/NoiseMol}{NoiseMol} \cite{jiang2023noisemol} & 2023/01 &  &  & {\Large $\bullet$} &  &  &  & \\
    & & \href{https://github.com/gaojianl/TransFoxMol}{TransFoxMol} \cite{gao2023transfoxmol} & 2023/01 &  &  & {\Large $\bullet$} &  &  &  &  \\
    \cline{2-4}
    & Graphormer \cite{ying2021transformers} & \href{https://github.com/shiokoo/CAFE-MPP}{CAFE-MPP} \cite{xie2023self} & 2023/01 &  &  & {\Large $\bullet$} &  &  &  & \\
    \hline
    \multirow{5}{*}{Decoder-only} & \multirow{3}{*}{GPT} 
    &  \href{https://github.com/devalab/molgpt}{MolGPT} \cite{bagal2021molgpt} & 2021/10 &   &  &  &  &  & {\Large \(\circ\)} & \\
    & & \href{https://github.com/paperswithcode/galai}{Galactica} \cite{taylor2022galactica} & 2022/12 &  &  & {\Large $\bullet$} &  &  &  & {\Large $\bullet$} \\
    & & MolXPT \cite{liu2023molxpt} & 2023/05  & {\Large $\bullet$} &  & {\Large $\bullet$} &  &  & {\Large $\bullet$} & \\
    \cline{2-4}
    & Transformer & \href{https://github.com/eyalmazuz/MolGen}{Taiga} \cite{mazuz2023molecule} & 2023/05 &  &  &  &  &  & {\Large \(\circ\)} & \\
    \cline{2-4}
    & \multirow{2}{*}{ChatGPT} & \href{https://github.com/phenixace/MolReGPT}{MolReGPT}  \cite{li2023empowering} & 2023/06  &  {\Large $\bullet$} &  &  &  &  & {\Large $\bullet$} & \\
    & & \href{https://github.com/ChnQ/LLM4Mol}{LLM4Mol} \cite{qian2023can} & 2023/07 &  &  & {\Large $\bullet$} &  &  &  & \\
    \hline
    \multirow{13}{*}{Text-Text} & \multirow{2}{*}{BART \cite{lewis2020bart}} & \href{https://github.com/MolecularAI/Chemformer}{Chemformer} \cite{irwin2022chemformer} & 2022/02 & &  &  &  &  & {\Large \(\circ\)} & \\
    & & \href{https://github.com/zjunlp/MolGen}{MolGen} \cite{fang2023domainagnostic} & 2023/05 &  &  &  &  &  & {\Large \(\circ\)} & \\
    \cline{2-4}
    & \multirow{8}{*}{T5} & \href{https://github.com/dhroth/c5t5}{C5T5} \cite{rothchild2021c5t5} & 2021/08 &  &  &  &  &  & {\Large \(\circ\)} & \\
    & & \href{https://github.com/HelloJocelynLu/t5chem}{T5Chem} \cite{Lu_Zhang_2022} & 2022/03 &  &  &  &  &  &  & {\Large $\bullet$} \\
    & & \href{https://github.com/blender-nlp/MolT5}{MolT5} \cite{edwards2022translation} & 2022/11 & 
    {\Large $\bullet$} &  &  &  &  & {\Large $\bullet$} & \\
    & & \href{https://github.com/GT4SD/multitask_text_and_chemistry_t5}{Text+Chem T5}\cite{christofidellis2023unifying} & 2023/05 &  {\Large $\bullet$} &  &  &  &  & {\Large $\bullet$} & {\Large $\bullet$} \\
    & & \href{https://github.com/Ellenzzn/ChatMol/tree/main}{ChatMol}  \cite{zeng2023interactive} & 2023/06 &  {\Large $\bullet$} &  &  &  &  & {\Large $\bullet$} & \\
    & & \href{https://github.com/AspirinCode/TransAntivirus}{TransAntivirus} \cite{mao2023transformer} & 2023/06 &  & {\Large $\bullet$} &  &  &  & {\Large \(\circ\)} & \\
    & & \href{https://github.com/QizhiPei/BioT5}{BioT5} \cite{pei2023biot5} & 2023/10 &   {\Large $\bullet$} &  & {\Large $\bullet$} &  &  & {\Large $\bullet$} & \\
    & & nach0 \cite{livne2023nach0} & 2023/11 &  {\Large $\bullet$} &  & {\Large $\bullet$} &  &  & {\Large $\bullet$} & {\Large $\bullet$} \\
    \cline{2-4}
    & \multirow{3}{*}{Transformer} & \href{https://app.syntelly.com/smiles2iupac}{Struct2IUPAC} \cite{krasnov2021struct2iupac} & 2020/11 &  & {\Large $\bullet$} &  &  &  &  & \\
    & & Transformer-R \cite{he2022transformer} & 2021/04 &   &  &  &  &  & {\Large \(\circ\)} & \\
    & & \href{https://github.com/ur-whitelab/chemcrow-public}{ChemCrow}\cite{bran2023chemcrow} & 2023/04 &  &  &  &  &  & {\Large \(\circ\)} & {\Large $\bullet$} \\
    \hline
    \multirow{5}{*}{Image-Text} & \multirow{2}{*}{Swin Transformer \cite{liu2021swin}} & \href{https://github.com/suanfaxiaohuo/SwinOCSR}{SwinOCSR} \cite{xu2022swinocsr} & 2022/07 &  &  &  &  & {\Large $\bullet$} &  & \\
    & & \href{https://github.com/thomas0809/MolScribe}{MolScribe} \cite{qian2023molscribe} & 2023/03 &  &  &  &  & {\Large $\bullet$} &  & \\
    \cline{2-4}
    & \multirow{3}{*}{Transformer} & \href{https://github.com/Kohulan/DECIMER-Image_Transformer}{DECIMER1.0} \cite{rajan2021decimer} & 2021/08 &  &  &  &  & {\Large $\bullet$} &  & \\
    & & \href{https://github.com/syntelly/img2smiles generator}{Image2SMILES}\cite{khokhlov2022image2smiles} & 2022/01 &  &  &  &  & {\Large $\bullet$} &  & \\
    & & \href{https://github.com/Jiacai-Yi/MICER}{MICER} \cite{yi2022micer} & 2022/08 &  &  &  &  & {\Large $\bullet$} &  & \\
    \hline
    \multirow{4}{*}{Graph-Text}    
    & \multirow{2}{*}{Sci-BERT}& \href{https://github.com/cnedwards/text2mol}{Text2Mol} \cite{edwards2021text2mol} & 2021/09 &  &  &  & {\Large $\bullet$} &  &  & \\
    & & \href{https://github.com/ddz16/MoMu}{MoMu} \cite{su2022molecular} & 2022/09 &  {\Large $\bullet$} &  & {\Large $\bullet$} & {\Large $\bullet$} &  & {\Large $\bullet$} & \\
    \cline{2-4}
    & MegaMolBART \cite{irwin2022chemformer} & \href{https://github.com/chao1224/MoleculeSTM}{MoleculeSTM} \cite{liu2023multi} & 2022/12 &  &  & {\Large $\bullet$} & {\Large $\bullet$} &  & {\Large \(\circ\)} & \\
    \cline{2-4}
    & BERT & \href{https://github.com/BioFM/OpenBioMed}{MolFM} \cite{luo2023molfm} & 2023/07 & {\Large $\bullet$} &  & {\Large $\bullet$} & {\Large $\bullet$} &  & {\Large $\bullet$} & \\
    \cline{2-4}
    & Q-Former \cite{li2023blip} & \href{https://github.com/eltociear/MolCA}{MolCA} \cite{liu2023molca} & 2023/11 & {\Large $\bullet$} & {\Large $\bullet$} & {\Large $\bullet$} & {\Large $\bullet$} &  &  & \\
    \cline{2-4}
    & KV-PLM & \href{https://github.com/gersteinlab/MolLM}{MolLM} \cite{tang2023mollm} & 2023/12 & {\Large $\bullet$} &  & {\Large $\bullet$} & {\Large $\bullet$} &  & {\Large \(\circ\)}& \\
    \hline
    \multirow{1}{*}{Image-Graph-Text} & Q-Former & \href{https://github.com/AI-HPC-Research-Team/GIT-Mol}{GIT-Mol} \cite{liu2023git} & 2023/08 & {\Large $\bullet$} &  & {\Large $\bullet$} &  & {\Large $\bullet$} & {\Large $\bullet$} & \\
    \hline
  \end{tabular}
  }
\end{table}

\section{Experimental results}

\textbf{SMILES Generation.}
Supplementary Table 3 illustrates the performance of SMILES generation across two methodologies: image recognition and text-based SMILES generation.
The image recognition task is an essential task for chemical image understanding.
The results show that models leveraging Swin Transformer (SwinOCSR) for encoding and t5 for decoding effectively generate SMILES from visual inputs.

\begin{table}[h]
\centering
\large
\caption*{\textbf{Supplementary Table 3:} Experimental results for SMILES generation}
\resizebox{\textwidth}{!}{%
\begin{tabular}{l|lllllllllll}
\hline
\textbf{Task} & \textbf{Input} & \textbf{Encoder} & \textbf{Decoder} & \textbf{Bleu $\uparrow$} & \textbf{Exact Match $\uparrow$} & \textbf{Levenshtein $\downarrow$} & \textbf{MACCS $\uparrow$} & \textbf{RDK $\uparrow$} & \textbf{Morgan $\uparrow$} & \textbf{Validity $\uparrow$} \\ \hline
\multirow{5}{*}{\shortstack{image recognition \\ (SMILES)}} & image & swinocsr & t5 & \textbf{0.885} & \textbf{0.356} & \textbf{12.746} & \textbf{0.951} & \textbf{0.890} & \textbf{0.868} & \textbf{0.908} \\
 & image & swinocsr & t511 & 0.833 & 0.175 & 18.589 & 0.897 & 0.785 & 0.765 & 0.761 \\
 & image & vit & t5 & 0.826 & 0.179 & 19.802 & 0.889 & 0.780 & 0.734 & 0.873 \\
 & image & swin & t5 & 0.782 & 0.067 & 24.726 & 0.794 & 0.633 & 0.581 & 0.891 \\
 & image & swin & t511 & 0.779 & 0.066 & 25.203 & 0.802 & 0.648 & 0.597 & 0.850 \\ \hline
\multirow{10}{*}{\shortstack{molecule generation \\ (text-based)}} & IUPAC & t5 & t5 & 
\textcolor{red}{\textbf{0.881}} & \textcolor{red}{\textbf{0.327}} & 14.209 & 0.909 & \textcolor{red}{\textbf{0.811}} & \textcolor{red}{\textbf{0.773}} & 0.905 \\
 & IUPAC & t511 & t511 & 0.881 & 0.323 & \textcolor{red}{\textbf{14.141}} & \textcolor{red}{\textbf{0.911}} & 0.808 & 0.771 & 0.893 \\
 & IUPAC & molt5 & molt5 & 0.853 & 0.206 & 17.645 & 0.881 & 0.758 & 0.723 & 0.773 \\
 & IUPAC & scibert & molt5 & 0.614 & 0.054 & 51.003 & 0.772 & 0.590 & 0.526 & 0.680 \\
 & IUPAC & bert & t5 & 0.296 & 0.000 & 96.484 & 0.409 & 0.268 & 0.128 & \textcolor{red}{\textbf{0.988}} \\ 
 \cline{2-11}
 & caption & molt5 & molt5 & \textbf{0.754} & 0.090 & \textbf{28.034} & 0.808 & 0.673 & 0.594 & 0.837 \\
 & caption & t5 & t5 & 0.691 & 0.097 & 34.891 & 0.814 & 0.677 & 0.600 & \textbf{0.858} \\
 & caption & t511 & t511 & 0.649 & \textbf{0.110} & 40.569 & \textbf{0.819} & \textbf{0.687} & \textbf{0.608} & 0.857 \\
 & caption & scibert & molt5 & 0.543 & 0.012 & 57.118 & 0.670 & 0.492 & 0.431 & 0.693 \\
 & caption & chemberta & molt5 & 0.425 & 0.002 & 73.061 & 0.561 & 0.408 & 0.325 & 0.740 \\ \hline
\end{tabular}%
}
\label{tab:SMILES Generation}
\end{table}

Within the text-based SMILES generation, the combination of IUPAC nomenclature with the t5 model emerges as the most effective, highlighting the model's ability to process and convert complex chemical language into accurate SMILES structures.
Moreover, the molt5 model's superior performance in generating molecules from captions suggests that its pre-training on SMILES expressions is advantageous for this specific task, emphasizing the value of domain-specific pre-training in enhancing model performance.

\begin{table}[h]
\centering
\caption*{\textbf{Supplementary Table 4:} Experimental results for IUPAC and caption generation}
\resizebox{\textwidth}{!}{%
\begin{tabular}{l|lllllllllll}
\hline
\textbf{Task} & \textbf{Input} & \textbf{Encoder} & \textbf{Decoder} & \textbf{Bleu-2 $\uparrow$} & \textbf{Bleu-4 $\uparrow$} & \textbf{Rouge-1 $\uparrow$} & \textbf{Rouge-2 $\uparrow$} & \textbf{Rouge-L $\uparrow$} & \textbf{Meteor-score $\uparrow$} & \textbf{Exact Match $\uparrow$} \\
\hline
\multirow{5}{*}{\shortstack{IUPAC \\ recognition \\(image)}} & image & swinocsr & t5 & \textbf{0.820} & \textbf{0.758} & \textbf{0.781} & \textbf{0.621} & \textbf{0.740} & \textbf{0.799} & \textbf{0.220} \\
 & image & swinocsr & molt5 & 0.803 & 0.730 & 0.751 & 0.573 & 0.707 & 0.777 & 0.158 \\
 & image & swinocsr & t511 & 0.800 & 0.730 & 0.757 & 0.577 & 0.711 & 0.779 & 0.170 \\
 & image & vit & t511 & 0.745 & 0.655 & 0.657 & 0.445 & 0.603 & 0.677 & 0.088 \\
 & image & vit & molt5 & 0.725 & 0.632 & 0.644 & 0.431 & 0.589 & 0.667 & 0.069 \\
\hline
\multirow{20}{*}{\shortstack{IUPAC \\ recognition}}
& SMILES & t5 & t5 & \textcolor{red}{\textbf{0.820}} & \textcolor{red}{\textbf{0.755}} & \textcolor{red}{\textbf{0.751}} & \textcolor{red}{\textbf{0.573}} & \textcolor{red}{\textbf{0.703}} & \textcolor{red}{\textbf{0.767}} & \textcolor{red}{\textbf{0.210}} \\
& SMILES & t511 & t511 & 0.794 & 0.719 & 0.711 & 0.515 & 0.660 & 0.732 & 0.146 \\
& SMILES & chemberta & molt5 & 0.786 & 0.710 & 0.717 & 0.524 & 0.669 & 0.739 & 0.134 \\
& SMILES & molt5 & molt5 & 0.752 & 0.665 & 0.669 & 0.459 & 0.615 & 0.688 & 0.095 \\
& SMILES & bart & bart & 0.598 & 0.500 & 0.684 & 0.463 & 0.624 & 0.676 & 0.170 \\
\cline{2-11}
& InChI & t5 & t5 & \textbf{0.744} & \textbf{0.658} & \textbf{0.648} & \textbf{0.442} & \textbf{0.591} & \textbf{0.663} & \textbf{0.131} \\
& InChI & t511 & t511 & 0.728 & 0.636 & 0.622 & 0.410 & 0.565 & 0.641 & 0.088 \\
& InChI & molt5 & molt5 & 0.728 & 0.635 & 0.622 & 0.413 & 0.566 & 0.642 & 0.093 \\
& InChI & chemberta & molt5 & 0.646 & 0.541 & 0.550 & 0.326 & 0.493 & 0.574 & 0.037 \\
& InChI & roberta & molt5 & 0.620 & 0.505 & 0.494 & 0.249 & 0.429 & 0.509 & 0.012 \\
\cline{2-11}
& SELFIES & molt5 & molt5 & \textbf{0.751} & \textbf{0.666} & 0.665 & 0.458 & 0.611 & 0.687 & 0.111 \\
& SELFIES & t5 & t5 & 0.744 & 0.658 & \textbf{0.668} & \textbf{0.465} & \textbf{0.615} & \textbf{0.692} & \textbf{0.132} \\
& SELFIES & t511 & t511 & 0.720 & 0.633 & 0.655 & 0.440 & 0.601 & 0.677 & 0.097 \\
& SELFIES & biogpt & biogpt & 0.669 & 0.585 & 0.594 & 0.391 & 0.534 & 0.602 & 0.005 \\
& SELFIES & bart & bart & 0.550 & 0.433 & 0.568 & 0.311 & 0.501 & 0.569 & 0.063 \\
\cline{2-11}
& graph & gcn & t511 & \textbf{0.719} & \textbf{0.643} & \textbf{0.640} & \textbf{0.436} & \textbf{0.591} & \textbf{0.667} & \textbf{0.056} \\
& graph & gat & t5 & 0.708 & 0.630 & 0.620 & 0.420 & 0.572 & 0.652 & 0.046 \\
& graph & gat & t511 & 0.685 & 0.603 & 0.597 & 0.393 & 0.550 & 0.628 & 0.037 \\
& graph & gin & t5 & 0.682 & 0.600 & 0.615 & 0.403 & 0.558 & 0.653 & 0.052 \\
& graph & gin & molt5 & 0.662 & 0.580 & 0.605 & 0.395 & 0.555 & 0.643 & 0.029 \\
\cline{2-11}
& caption & t511 & t511 & \textbf{0.736} & \textbf{0.665} & \textbf{0.707} & \textbf{0.532} & 0.660 & \textbf{0.716} & \textbf{0.185} \\
& caption & biogpt & biogpt & 0.718 & 0.655 & 0.650 & 0.479 & 0.605 & 0.673 & 0.000 \\
& caption & molt5 & molt5 & 0.699 & 0.615 & 0.652 & 0.460 & 0.604 & 0.667 & 0.129 \\
& caption & t5 & t5 & 0.621 & 0.545 & 0.675 & 0.495 & \textbf{0.627} & 0.679 & 0.183 \\
& caption & scibert & molt5 & 0.596 & 0.503 & 0.531 & 0.314 & 0.478 & 0.568 & 0.025 \\
\hline
\multirow{25}{*}{\shortstack{molecule \\ captioning}}
& IUPAC & t5 & t5 & \textcolor{red}{\textbf{0.547}} & \textcolor{red}{\textbf{0.457}} & \textcolor{red}{\textbf{0.629}} & \textcolor{red}{\textbf{0.476}} & \textcolor{red}{\textbf{0.566}} & \textcolor{red}{\textbf{0.579}} & /\\
& IUPAC & t511 & t511 & 0.535 & 0.446 & 0.622 & 0.469 & 0.559 & 0.571 & /\\
& IUPAC & molt5 & molt5 & 0.525 & 0.433 & 0.612 & 0.455 & 0.549 & 0.556 & /  \\
& IUPAC & bart & bart & 0.455 & 0.352 & 0.568 & 0.391 & 0.489 & 0.484 & /  \\
& IUPAC & chemberta & molt5 & 0.424 & 0.333 & 0.550 & 0.386 & 0.492 & 0.490 & / \\
\cline{2-11}
& SMILES & molt5 & molt5 & \textbf{0.535} & \textbf{0.436} & \textbf{0.625} & \textbf{0.466} & \textbf{0.559} & \textbf{0.563} & / \\
& SMILES & t5 & t5 & 0.508 & 0.414 & 0.605 & 0.446 &0.545 & 0.543 & / \\
& SMILES & t511 & t511 & 0.488 & 0.392 & 0.586 & 0.426 & 0.528 & 0.522 & /  \\
& SMILES & chemberta & molt5 & 0.488 & 0.388 & 0.578 & 0.417 & 0.522 & 0.520  & / \\
& SMILES & bart & bart & 0.406 & 0.301 & 0.538 & 0.359 & 0.465 & 0.443 & / \\
\cline{2-11}
& InChI & t5 & t5 & \textbf{0.477} & \textbf{0.383} & \textbf{0.577} & \textbf{0.415} & \textbf{0.518} & \textbf{0.511}  & /\\
& InChI & t511 & t511 & 0.464 & 0.369 & 0.567 & 0.404 & 0.510 & 0.499  & /\\
& InChI & molt5 & molt5 & 0.461 & 0.367 & 0.571 & 0.406 & 0.513 & 0.500 & / \\
& InChI & chemberta & molt5 & 0.451 & 0.356 & 0.557 & 0.394 & 0.502 & 0.490  & /\\
& InChI & bart & bart & 0.393 & 0.284 & 0.519 & 0.335 & 0.447 & 0.429  & /\\
\cline{2-11}
& SELFIES & t5 & t5 & \textbf{0.462} & \textbf{0.360} & \textbf{0.569} & \textbf{0.400} & \textbf{0.506} & \textbf{0.503} & / \\
& SELFIES & t511 & t511 & 0.446 & 0.348 & 0.559 & 0.392 & 0.499 & 0.490 & / \\
& SELFIES & chemberta & molt5 & 0.439 & 0.340 & 0.549 & 0.381 & 0.491 & 0.478 & / \\
& SELFIES & bart & bart & 0.377 & 0.266 & 0.510 & 0.320 & 0.434 & 0.414 & / \\
& SELFIES & biogpt & biogpt & 0.277 & 0.204 & 0.425 & 0.250 & 0.355 & 0.345 & / \\
\cline{2-11}
& graph & gat & molt5 & \textbf{0.419} & \textbf{0.318} & \textbf{0.521} & \textbf{0.345} & \textbf{0.460} & \textbf{0.449} & / \\
& graph & gin & molt5 & 0.407 & 0.305 & 0.512 & 0.332 & 0.450 & 0.437 & / \\
& graph & gcn & molt5 & 0.392 & 0.291 & 0.512 & 0.333 & 0.450 & 0.430 & / \\
& graph & gin & t511 & 0.362 & 0.256 & 0.479 & 0.295 & 0.418 & 0.401 & / \\
& graph & gcn & bart & 0.358 & 0.251 & 0.489 & 0.298 & 0.413 & 0.391 & / \\
\cline{2-11}
& image & swinocsr & molt5 & \textbf{0.495} & \textbf{0.399} & \textbf{0.590} & \textbf{0.429} & \textbf{0.532} & \textbf{0.527} & / \\
& image & swinocsr & t511 & 0.492 & 0.394 & 0.585 & 0.422 & 0.527 & 0.524 & / \\
& image & swinocsr & t5 & 0.417 & 0.317 & 0.533 & 0.361 & 0.472 & 0.461 & / \\
& image & swin & molt5 & 0.409 & 0.307 & 0.517 & 0.343 & 0.460 & 0.442 & / \\
& image & swin & t511 & 0.400 & 0.301 & 0.517 & 0.344 & 0.461 & 0.435 & / \\
\hline
\end{tabular}%
}
\label{tab:IUPAC Generation}
\end{table}

\textbf{IUPAC Generation.}
Supplementary Table 4 emphasizes the tasks associated with IUPAC Generation, focusing on the recognition of IUPAC names from molecular images and internal information.
Our benchmark does not include experiments for generating IUPAC names from captions because some captions already contain IUPAC names.

\begin{table}[ht]
\centering
\caption*{\textbf{Supplementary Table 5:} Experimental results for retrieval}
\resizebox{\textwidth}{!}{%
\begin{tabular}{l|llllllll}
\hline
\textbf{Task} & \textbf{Input} & \textbf{Encoder} & \textbf{Pooling} & \textbf{Target Modality} & \textbf{R@1 $\uparrow$} & \textbf{MRR $\uparrow$} & \textbf{R@5 $\uparrow$} & \textbf{R@10 $\uparrow$}\\
\hline
\multirow{25}{*}{\shortstack{molecular retrieval\\ (caption)}} 
& IUPAC & t5 & max & caption & $ \textcolor{red}{\textbf{0.637}}\pm0.024$ & $ \textcolor{red}{\textbf{0.741}}\pm0.022$ & $ \textcolor{red}{\textbf{0.880}}\pm0.019$ & $ \textcolor{red}{\textbf{0.936}}\pm0.012 $ \\
& IUPAC & scibert & avg & caption & $ 0.635\pm0.084$ & $ 0.736\pm0.076$ & $ 0.870\pm0.063$ & $ 0.935\pm0.051 $ \\
& IUPAC & scibert & max & caption & $ 0.581\pm0.061$ & $ 0.694\pm0.056$ & $ 0.848\pm0.046$ & $ 0.918\pm0.036 $ \\
& IUPAC & molt5 & avg & caption & $ 0.578\pm0.026$ & $ 0.701\pm0.026$ & $ 0.869\pm0.027$ & $ 0.931\pm0.016 $ \\
& IUPAC & t5 & avg & caption & $ 0.564\pm0.035$ & $ 0.685\pm0.034$ & $ 0.855\pm0.032$ & $ 0.923\pm0.022 $ \\
\cline{2-9}
& SMILES & bart & avg & caption & $ \textbf{0.440}\pm0.043$ & $ \textbf{0.589}\pm0.043$ & $ 0.797\pm0.044$ & $ 0.895\pm0.033 $ \\
& SMILES & t5 & max & caption & $ 0.437\pm0.046$ & $ 0.588\pm0.046$ & $ \textbf{0.800}\pm0.045$ & $ 0.892\pm0.028 $ \\
& SMILES & t5 & avg & caption & $ 0.427\pm0.056$ & $ 0.583\pm0.056$ & $ 0.797\pm0.056$ & $ \textbf{0.897}\pm0.037 $ \\
& SMILES & roberta & max & caption & $ 0.426\pm0.070$ & $ 0.572\pm0.075$ & $ 0.772\pm0.082$ & $ 0.880\pm0.066 $ \\
& SMILES & molt5 & avg & caption & $ 0.402\pm0.054$ & $ 0.557\pm0.055$ & $ 0.772\pm0.059$ & $ 0.875\pm0.037 $ \\
\cline{2-9}
& InChI & t5 & avg & caption & $ \textbf{0.339}\pm0.033$ & $ \textbf{0.504}\pm0.036$ & $ \textbf{0.733}\pm0.040$ & $ \textbf{0.858}\pm0.030 $ \\
& InChI & scibert & avg & caption & $ 0.337\pm0.076$ & $ 0.498\pm0.095$ & $ 0.719\pm0.121$ & $ 0.843\pm0.111 $ \\
& InChI & t511 & avg & caption & $ 0.325\pm0.053$ & $ 0.488\pm0.062$ & $ 0.717\pm0.077$ & $ 0.846\pm0.062 $ \\
& InChI & bart & avg & caption & $ 0.289\pm0.036$ & $ 0.444\pm0.045$ & $ 0.661\pm0.059$ & $ 0.808\pm0.051 $ \\
& InChI & molt5 & avg & caption & $ 0.280\pm0.055$ & $ 0.433\pm0.064$ & $ 0.646\pm0.079$ & $ 0.790\pm0.068 $ \\
\cline{2-9}
& SELFIES & roberta & max & caption & $ \textbf{0.436}\pm0.079$ & $ \textbf{0.579}\pm0.081$ & $ \textbf{0.778}\pm0.086$ & $ \textbf{0.875}\pm0.069 $ \\
& SELFIES & scibert & max & caption & $ 0.406\pm0.053$ & $ 0.553\pm0.057$ & $ 0.757\pm0.063$ & $ \textbf{0.875}\pm0.051 $ \\
& SELFIES & bert & max & caption & $ 0.393\pm0.055$ & $ 0.546\pm0.063$ & $ 0.751\pm0.073$ & $ 0.868\pm0.062 $ \\
& SELFIES & bart & avg & caption & $ 0.367\pm0.053$ & $ 0.520\pm0.057$ & $ 0.731\pm0.065$ & $ 0.853\pm0.055 $ \\
& SELFIES & molt5 & max & caption & $ 0.339\pm0.041$ & $ 0.492\pm0.045$ & $ 0.709\pm0.054$ & $ 0.835\pm0.045 $ \\
\cline{2-9}
& graph & gat & max & caption & $ \textbf{0.259}\pm0.060$ & $ \textbf{0.406}\pm0.079$ & $ \textbf{0.610}\pm0.111$ & $ \textbf{0.765}\pm0.111 $ \\
& graph & gcn & max & caption & $ 0.242\pm0.048$ & $ 0.383\pm0.059$ & $ 0.576\pm0.077$ & $ 0.724\pm0.068 $ \\
& graph & gin & max & caption & $ 0.241\pm0.030$ & $ 0.384\pm0.040$ & $ 0.577\pm0.053$ & $ 0.732\pm0.053 $ \\
& graph & gat & avg & caption & $ 0.189\pm0.046$ & $ 0.319\pm0.068$ & $ 0.501\pm0.102$ & $ 0.648\pm0.108 $ \\
& graph & gin & avg & caption & $ 0.185\pm0.038$ & $ 0.309\pm0.058$ & $ 0.476\pm0.086$ & $ 0.630\pm0.099 $ \\
\cline{2-9}
& image & swinocsr & max & caption & $ \textbf{0.411}\pm0.124$ & $ \textbf{0.552}\pm0.146$ & $ \textbf{0.748}\pm0.181$ & $ 0.844\pm0.164 $ \\
& image & swinocsr & avg & caption & $ 0.384\pm0.094$ & $ 0.534\pm0.106$ & $ 0.747\pm0.127$ & $ \textbf{0.856}\pm0.107 $ \\
& image & resnet & avg & caption & $ 0.233\pm0.041$ & $ 0.385\pm0.055$ & $ 0.587\pm0.075$ & $ 0.753\pm0.073 $ \\
& image & vit & max & caption & $ 0.164\pm0.045$ & $ 0.285\pm0.068$ & $ 0.443\pm0.099$ & $ 0.607\pm0.116 $ \\
& image & vit & avg & caption & $ 0.121\pm0.019$ & $ 0.239\pm0.034$ & $ 0.399\pm0.056$ & $ 0.574\pm0.070 $ \\
\hline
\multirow{25}{*}{\shortstack{molecular retrieval\\ (IUPAC)}} 
& SMILES & t5 & avg & IUPAC & $ \textcolor{red}{\textbf{0.709}}\pm0.069$ & $ \textcolor{red}{\textbf{0.817}}\pm0.054$ & $ \textcolor{red}{\textbf{0.958}}\pm0.033$ & $ \textcolor{red}{\textbf{0.986}}\pm0.015 $ \\
& SMILES & t511 & max & IUPAC & $ 0.698\pm0.116$ & $ 0.802\pm0.096$ & $ 0.941\pm0.065$ & $ 0.975\pm0.032 $ \\
& SMILES & scibert & avg & IUPAC & $ 0.681\pm0.147$ & $ 0.789\pm0.127$ & $ 0.930\pm0.097$ & $ 0.967\pm0.060 $ \\
& SMILES & t511 & avg & IUPAC & $ 0.655\pm0.099$ & $ 0.775\pm0.082$ & $ 0.930\pm0.055$ & $ 0.970\pm0.028 $ \\
& SMILES & bart & max & IUPAC & $ 0.629\pm0.090$ & $ 0.759\pm0.075$ & $ 0.935\pm0.051$ & $ 0.978\pm0.025 $ \\
\cline{2-9}
& InChI & t5 & avg & IUPAC & $ \textbf{0.545}\pm0.064$ & $ \textbf{0.687}\pm0.059$ & $ 0.877\pm0.054$ & $ 0.945\pm0.028 $ \\
& InChI & t5 & max & IUPAC & $ 0.519\pm0.074$ & $ 0.661\pm0.065$ & $ 0.850\pm0.051$ & $ 0.931\pm0.031 $ \\
& InChI & t511 & avg & IUPAC & $ 0.518\pm0.114$ & $ 0.665\pm0.111$ & $ 0.867\pm0.106$ & $ 0.936\pm0.073 $ \\
& InChI & bart & avg & IUPAC & $ 0.502\pm0.085$ & $ 0.662\pm0.084$ & $ \textbf{0.882}\pm0.085$ & $ \textbf{0.950}\pm0.049 $ \\
& InChI & t511 & max & IUPAC & $ 0.499\pm0.099$ & $ 0.654\pm0.094$ & $ 0.868\pm0.090$ & $ 0.945\pm0.046 $ \\
\cline{2-9}
& SELFIES & t511 & max & IUPAC & $ \textbf{0.626}\pm0.068$ & $ \textbf{0.753}\pm0.057$ & $ \textbf{0.920}\pm0.038$ & $ \textbf{0.972}\pm0.021 $ \\
& SELFIES & bart & avg & IUPAC & $ 0.603\pm0.094$ & $ 0.731\pm0.083$ & $ 0.906\pm0.067$ & $ 0.962\pm0.038 $ \\
& SELFIES & t5 & avg & IUPAC & $ 0.570\pm0.076$ & $ 0.706\pm0.069$ & $ 0.887\pm0.059$ & $ 0.950\pm0.035 $ \\
& SELFIES & bart & max & IUPAC & $ 0.508\pm0.095$ & $ 0.662\pm0.094$ & $ 0.876\pm0.097$ & $ 0.947\pm0.063 $ \\
& SELFIES & molt5 & avg & IUPAC & $ 0.506\pm0.112$ & $ 0.643\pm0.112$ & $ 0.828\pm0.114$ & $ 0.912\pm0.080 $ \\
\cline{2-9}
& graph & gcn & max & IUPAC & $ \textbf{0.279}\pm0.067$ & $ 0.421\pm0.084$ & $ 0.621\pm0.113$ & $ 0.756\pm0.107 $ \\
& graph & gin & max & IUPAC & $ 0.272\pm0.068$ & $ \textbf{0.423}\pm0.088$ & $ \textbf{0.631}\pm0.118$ & $ \textbf{0.773}\pm0.114 $ \\
& graph & gin & avg & IUPAC & $ 0.216\pm0.071$ & $ 0.352\pm0.104$ & $ 0.536\pm0.152$ & $ 0.684\pm0.169 $ \\
& graph & gat & avg & IUPAC & $ 0.204\pm0.074$ & $ 0.336\pm0.112$ & $ 0.518\pm0.170$ & $ 0.669\pm0.191 $ \\
& graph & gcn & avg & IUPAC & $ 0.186\pm0.030$ & $ 0.321\pm0.046$ & $ 0.500\pm0.068$ & $ 0.661\pm0.080 $ \\
\cline{2-9}
& image & swin & max & IUPAC & $ \textbf{0.696}\pm0.083$ & $ \textbf{0.803}\pm0.069$ & $ \textbf{0.945}\pm0.050$ & $ \textbf{0.979}\pm0.030 $ \\
& image & swin & avg & IUPAC & $ 0.462\pm0.104$ & $ 0.608\pm0.105$ & $ 0.808\pm0.104$ & $ 0.900\pm0.076 $ \\
& image & resnet & avg & IUPAC & $ 0.289\pm0.057$ & $ 0.450\pm0.070$ & $ 0.674\pm0.091$ & $ 0.834\pm0.080 $ \\
& image & resnet & max & IUPAC & $ 0.215\pm0.070$ & $ 0.376\pm0.110$ & $ 0.598\pm0.170$ & $ 0.761\pm0.188 $ \\
& image & vit & max & IUPAC & $ 0.215\pm0.080$ & $ 0.365\pm0.117$ & $ 0.575\pm0.172$ & $ 0.740\pm0.178 $ \\
\hline
\end{tabular}%
}
\label{tab:retrieval}
\end{table}

For IUPAC name recognition from the molecular images, the combination of swinocsr and t5 models yields the best performance, similar to the SMILES generation task.
It suggests that the swinocsr, originally pre-trained on molecular SMILES and images, is also adept at transfer learning for IUPAC name recognition tasks.
On the other hand, combining SMILES data with the t5 model achieves the best results for IUPAC recognition based on molecular internal information.
Although graph-based models show only slightly lower performance in textual similarity metrics compared to language models, they exhibit a relative shortfall in achieving exact matches.

\textbf{Molecule Captioning.}
The molecular captions encapsulate the essence of the molecule along with its biochemical and physicochemical properties.
Molecule captioning involves the conversion of various representation forms into captions.
As shown in Supplementary Table 4, unlike image recognition tasks for SMILES and IUPAC names, experiments on caption generation show that performance based on IUPAC and SMILES outperforms those based on image-derived captions.
It suggests that image encoders excel at distilling key points from images.  However, as the diversity of generated content increases and the length of the generated sequences extends, the performance of image encoder models tends to fall short of language models.
Within the scope of language models, text description generation based on IUPAC is superior, partly because the target captions commonly include IUPAC names.  Additionally, graph encoders exhibit weaker performance in text generation tasks, further emphasizing the strengths of language models in handling complex linguistic structures and content variability.

\textbf{Molecular Retrieval.}
In molecular cross-modal retrieval tasks, the focus is primarily on the retrieval between molecular external information and either internal or external information.
We categorize the retrieval tasks into two main types, targeting captions and IUPAC, further subdivided into nine specific retrieval tasks.
As shown in Supplementary Table 5, the IUPAC-to-caption and SMILES-to-IUPAC tasks achieve the best results in their respective categories.
Regarding the models, the t5 family demonstrates superiority over bart, encoder-only models, and graph encoders.

\textbf{Property Prediction.}
\begin{table}[h]
\centering
\caption*{\textbf{Supplementary Table 6:} Comparative performance of traditional machine learning (ML), GNN, and Transformer models on property prediction classification tasks.
The evaluation metric is AUC\_ROC.
The numbers listed under each dataset represent the count of molecules and the number of classification tasks.
The (G) denotes graph-based inputs, while (S) denotes SMILES-based inputs as the input modalities.}
\label{tab:tml_g_t_c}
\resizebox{\textwidth}{!}{%
\begin{tabular}{@{}llccccccc@{}}
\toprule
\multirow{3}{*}{Method} & \multirow{3}{*}{Model}       & BBBP  & Tox21 & ToxCast & SIDER & ClinTox & BACE  & Avg   \\ 
&    & 2039  & 7831  & 8575    & 1427  & 1478   & 1513  \\ 
&    & 1     & 12    & 617     & 27    & 2      & 1     \\
\toprule
\multirow{3}{*}{Traditional ML Models} 
& Logistic Regression \cite{wu2018moleculenet}  & 0.699 & 0.664 & 0.600   & 0.596 & 0.782  & 0.781 & 0.687 \\
& KernelSVM \cite{cortes1995support} & 0.729 & 0.679 & 0.652   & 0.620 & 0.810  & 0.862 & 0.725 \\
& RF \cite{breiman2001random} & 0.714 & 0.653 & 0.634   & 0.633 & 0.765  & 0.867 & 0.711 \\
\midrule
\multirow{5}{*}{GNN-based}
& GIN \cite{xu2018powerful}  & 0.640 & 0.701 & 0.613   & 0.550 & 0.580  & 0.700 & 0.631 \\
& GraphCL \cite{you2020graph}         & 0.675 & 0.750 & 0.628   & 0.601 & 0.789  & 0.687 & 0.688 \\
& GROVER \cite{rong2020self}         & 0.700 & 0.743 & 0.654   & 0.648 & 0.812  & 0.826 & 0.731 \\
& GraphMVP \cite{liu2022pretraining}       & 0.724 & 0.744 & 0.631   & 0.639 & 0.775  & 0.812 & 0.721 \\
& MoMu-SciBERT(G) \cite{su2022molecular}  & 0.705 & 0.756 & 0.634   & 0.605 & 0.799  & 0.767 & 0.711 \\ \midrule
\multirow{3}{*}{Transformer-based}    
& MolT5 \cite{edwards2022translation}         & 0.638 & 0.736 & 0.556   & 0.526 & 0.909  & 0.708 & 0.679 \\
& KV-PLM \cite{zeng2022deep}        & 0.669 & 0.647 & 0.586   & 0.553 & 0.843  & 0.719 & 0.670 \\
& MoleculeSTM(S) \cite{liu2023multi}  & 0.708 & 0.757 & 0.652   & 0.637 & 0.866  & 0.820 & 0.740 \\ \bottomrule
\end{tabular}%
}
\end{table}
We categorize the tasks into classification and regression, testing them across scaffold and random data splitting methodologies.
The evaluation extends to nine distinct datasets, each encompassing three text modalities, eight text encoders, and three graph models \cite{xu2018powerful}\cite{kipf2017semisupervised}\cite{veličković2018graph}, integrated with two different pooling strategies.

We have selected several commonly used traditional machine learning methods from MoleculeNet, as well as typical GNN-based and transformer-based models.
The results in Supplementary Table 6 and Table 7 allow us to analyze the performance differences between traditional methods, graph models, and transformer-based models over recent years.
Each type of model excels in specific tasks.
However, both GNN-based and transformer-based models can enhance their performance through self-supervised pre-training strategies.
Furthermore, transformer-based models, with their larger parameter count, possess a higher potential for capturing complex patterns.
Consequently, each dataset is tested with 108 unique experimental setups.
For each combination of input data and model, we select the top five performing models, as shown in Supplementary Table 8 and Table 9.

\begin{table}[h]
\centering
\caption*{\textbf{Supplementary Table 7:} Comparative performance of traditional machine learning, GNN, and Transformer models on property prediction regression tasks.
The evaluation metric is RMSE.
The numbers provided under each dataset indicate the count of molecules.
The (G) denotes graph-based inputs, while (S) denotes SMILES-based inputs as the input modalities.}
\label{tab:tml_g_t_r}
\begin{tabular}{@{}llccc@{}}
\toprule
\multirow{2}{*}{Method} & \multirow{2}{*}{Model}       & ESOL  & Freesolv & Lipophilicity \\ 
& & 1128& 642&4200\\
\midrule
\multirow{3}{*}{Traditional ML Models} 
& Logistic Regression \cite{wu2018moleculenet}      & 0.732 & 0.973    & 1.003        \\
& KernelSVM \cite{cortes1995support}     & 0.752 & 1.086    & 0.718       \\
& RF \cite{breiman2001random}           & 0.817 & 1.336    & 0.791        \\
\midrule
\multirow{4}{*}{GNN-based}             
& GIN \cite{xu2018powerful}          & 1.642 & 3.781    & 0.821        \\
& GROVER \cite{rong2020self}        & 0.992 & 2.470    & 0.810        \\
& GraphMVP \cite{liu2022pretraining}     & 1.341 & 2.972    & 0.769        \\
& Uni-mol \cite{zhou2023unimol}      & 0.788 & 1.480    & 0.603        \\ \midrule
\multirow{2}{*}{Transformer-based}     
& SMILES-TSFM \cite{honda2019smiles}   & 1.339 & 3.097    & 1.215        \\
& MolT5 \cite{edwards2022translation}        & 1.252 & 3.139    & 0.940        \\ 
\bottomrule
\end{tabular}
\end{table}

\begin{table}[ht!]
\centering
\caption*{\textbf{Supplementary Table 8:} Experimental results for classification tasks}
\resizebox{\textwidth}{!}{%
\begin{tabular}{l|l|lllllll}
\hline
\textbf{Task} & \textbf{Data Spliting} & \textbf{Dataset} & \textbf{Input} & \textbf{Encoder}& \textbf{Pooling}  & \textbf{roc\_auc $\uparrow$} & \textbf{pr\_auc $\uparrow$} & \textbf{f1\_score $\uparrow$} \\
\hline
\multirow{60}{*}{\shortstack{\textbf{property prediction} \\ (classification)}} &
\multirow{30}{*}{scaffold} 
& bace & SELFIES & roberta & max & $ \textbf{0.742}\pm0.036 $ & $ \textbf{0.750}\pm0.013  $ & $ 0.544\pm0.126 $ \\
& & bace & SELFIES & molt5 & max & $ 0.708\pm0.065  $ & $ 0.678\pm0.043  $ & $ 0.391\pm0.033 $ \\
& & bace & InChI & bart & avg & $ 0.704\pm0.058  $ & $ 0.699\pm0.063  $ & $ 0.516\pm0.112 $ \\
& & bace & InChI & t5 & max & $ 0.703\pm0.030  $ & $ 0.699\pm0.043  $ & $ \textbf{0.646}\pm0.032 $ \\
& & bace & InChI & molt5 & max & $ 0.678\pm0.039  $ & $ 0.699\pm0.036  $ & $ 0.565\pm0.038 $ \\
\cline{3-9}
& & bbbp & graph & gin & avg & $ \textbf{0.652}\pm0.009  $ & $ 0.684\pm0.024  $ & $ 0.528\pm0.026 $ \\
& & bbbp & graph & gat & avg & $ 0.651\pm0.013  $ & $ 0.691\pm0.013  $ & $ 0.470\pm0.057 $ \\
& & bbbp & SMILES & molt5 & avg & $ 0.651\pm0.010  $ & $ 0.692\pm0.010  $ & $ \textbf{0.602}\pm0.034 $ \\
& & bbbp & graph & gcn & max & $ 0.645\pm0.011  $ & $ \textbf{0.693}\pm0.013  $ & $ 0.492\pm0.118 $ \\
& & bbbp & graph & gcn & avg & $ 0.641\pm0.012  $ & $ 0.675\pm0.014  $ & $ 0.536\pm0.057 $ \\
\cline{3-9}
& & clintox & SELFIES & t511 & avg & $ \textbf{0.923}\pm0.003  $ & $ 0.722\pm0.047  $ & $ \textbf{0.483}\pm0.000 $ \\
& & clintox & InChI & t5 & avg & $ 0.908\pm0.011  $ & $ 0.700\pm0.022  $ & $ 0.483\pm0.000 $ \\
& & clintox & SELFIES & t5 & avg & $ 0.899\pm0.064  $ & $ \textbf{0.749}\pm0.096  $ & $ \textbf{0.483}\pm0.000 $ \\
& & clintox & SELFIES & t5 & max & $ 0.874\pm0.020  $ & $ 0.721\pm0.034  $ & $ 0.381\pm0.144 $ \\
& & clintox & InChI & molt5 & avg & $ 0.874\pm0.039  $ & $ 0.667\pm0.039  $ & $ \textbf{0.483}\pm0.000 $ \\
\cline{3-9}
& & sider & SMILES & t511 & avg & $ \textbf{0.563}\pm0.029  $ & $ 0.606\pm0.006  $ & $ 0.426\pm0.026 $ \\
& & sider & graph & gat & avg & $ 0.557\pm0.007  $ & $ 0.605\pm0.003  $ & $ 0.442\pm0.014 $ \\
& & sider & graph & gat & max & $ 0.551\pm0.021  $ & $ 0.598\pm0.014  $ & $ 0.455\pm0.010 $ \\
& & sider & InChI & t5 & avg & $ 0.550\pm0.016  $ & $ 0.601\pm0.014  $ & $ \textbf{0.459}\pm0.011 $ \\
& & sider & graph & gcn & max & $ 0.545\pm0.026  $ & $ \textbf{0.607}\pm0.016  $ & $ 0.443\pm0.003 $ \\
\cline{3-9}
& & tox21 & graph & gin & avg & $ \textbf{0.693}\pm0.019  $ & $ \textbf{0.248}\pm0.037  $ & $ 0.473\pm0.000 $ \\
& & tox21 & graph & gcn & avg & $ 0.679\pm0.007  $ & $ 0.231\pm0.007  $ & $ 0.473\pm0.000 $ \\
& & tox21 & SMILES & molt5 & avg & $ 0.664\pm0.031  $ & $ 0.200\pm0.058  $ & $ \textbf{0.488}\pm0.021 $ \\
& & tox21 & graph & gat & avg & $ 0.664\pm0.014  $ & $ 0.212\pm0.003  $ & $ 0.473\pm0.000 $ \\
& & tox21 & SMILES & t511 & avg & $ 0.657\pm0.043  $ & $ 0.212\pm0.050  $ & $ 0.486\pm0.018 $ \\
\cline{3-9}
& & toxcast & graph & gat & avg & $ \textbf{0.606}\pm0.005  $ & $ 0.305\pm0.002  $ & $ 0.449\pm0.003 $ \\
& & toxcast & graph & gcn & avg & $ 0.604\pm0.006  $ & $ \textbf{0.306}\pm0.002  $ & $ 0.452\pm0.003 $ \\
& & toxcast & graph & gin & avg & $ 0.599\pm0.008  $ & $ 0.294\pm0.016  $ & $ 0.446\pm0.001 $ \\
& & toxcast & SMILES & t511 & avg & $ 0.582\pm0.008  $ & $ 0.302\pm0.006  $ & $ 0.451\pm0.007 $ \\
& & toxcast & SELFIES & t511 & avg & $ 0.579\pm0.002  $ & $ 0.293\pm0.002  $ & $ \textbf{0.453}\pm0.002 $ \\
\cline{2-9}
&\multirow{30}{*}{random} 
& bace & graph & gcn & avg & $ \textbf{0.824}\pm0.034  $ & $ \textbf{0.752}\pm0.023  $ & $ \textbf{0.683}\pm0.142 $ \\
& & bace & graph & gat & avg & $ 0.782\pm0.022  $ & $ 0.715\pm0.037  $ & $ 0.398\pm0.039 $ \\
& & bace & graph & gin & avg & $ 0.751\pm0.034  $ & $ 0.684\pm0.061  $ & $ 0.399\pm0.042 $ \\
& & bace & SMILES & t5 & avg & $ 0.712\pm0.043  $ & $ 0.609\pm0.056  $ & $ 0.429\pm0.089 $ \\
& & bace & InChI & t511 & avg & $ 0.710\pm0.036  $ & $ 0.656\pm0.035  $ & $ 0.459\pm0.065 $ \\
\cline{3-9}
& & bbbp & graph & gcn & avg & $ \textbf{0.872}\pm0.014  $ & $ 0.935\pm0.017  $ & $ \textbf{0.769}\pm0.031 $ \\
& & bbbp & SMILES & molt5 & avg & $ 0.849\pm0.062  $ & $ 0.928\pm0.036  $ & $ 0.757\pm0.071 $ \\
& & bbbp & graph & gat & avg & $ 0.848\pm0.020  $ & $ 0.932\pm0.013  $ & $ 0.768\pm0.012 $ \\
& & bbbp & SMILES & t511 & avg & $ 0.848\pm0.015  $ & $ \textbf{0.940}\pm0.002  $ & $ 0.742\pm0.016 $ \\
& & bbbp & SMILES & t5 & avg & $ 0.847\pm0.027  $ & $ 0.932\pm0.017  $ & $ 0.756\pm0.003 $ \\
\cline{3-9}
& & clintox & SMILES & t5 & avg & $ \textbf{0.861}\pm0.041  $ & $ \textbf{0.654}\pm0.029  $ & $ \textbf{0.485}\pm0.000 $ \\
& & clintox & SMILES & molt5 & avg & $ 0.852\pm0.023  $ & $ 0.643\pm0.026  $ & $ 0.478\pm0.004 $ \\
& & clintox & SELFIES & t5 & avg & $ 0.850\pm0.004  $ & $ 0.611\pm0.027  $ & $ 0.484\pm0.003 $ \\
& & clintox & SMILES & t511 & avg & $ 0.845\pm0.063  $ & $ 0.652\pm0.088  $ & $ 0.482\pm0.009 $ \\
& & clintox & SELFIES & t511 & avg & $ 0.839\pm0.022  $ & $ 0.607\pm0.009  $ & $ 0.483\pm0.005 $ \\
\cline{3-9}
& & sider & SMILES & t5 & avg & $ \textbf{0.581}\pm0.031  $ & $ \textbf{0.629}\pm0.015  $ & $ \textbf{0.475}\pm0.017 $ \\
& & sider & graph & gat & max & $ 0.580\pm0.018  $ & $ 0.600\pm0.002  $ & $ 0.439\pm0.013 $ \\
& & sider & InChI & t5 & avg & $ 0.568\pm0.038  $ & $ 0.618\pm0.017  $ & $ 0.461\pm0.015 $ \\
& & sider & SMILES & molt5 & avg & $ 0.562\pm0.008  $ & $ 0.600\pm0.008  $ & $ 0.461\pm0.012 $ \\
& & sider & graph & gin & max & $ 0.561\pm0.018  $ & $ 0.618\pm0.011  $ & $ 0.451\pm0.008 $ \\
\cline{3-9}
& & tox21 & graph & gcn & avg & $ \textbf{0.795}\pm0.023  $ & $ \textbf{0.283}\pm0.021  $ & $ 0.481\pm0.001 $ \\
& & tox21 & graph & gin & avg & $ 0.752\pm0.025  $ & $ 0.224\pm0.025  $ & $ 0.480\pm0.001 $ \\
& & tox21 & SMILES & t511 & avg & $ 0.745\pm0.008  $ & $ 0.246\pm0.041  $ & $ \textbf{0.494}\pm0.021 $ \\
& & tox21 & SMILES & molt5 & avg & $ 0.738\pm0.011  $ & $ 0.268\pm0.011  $ & $ 0.492\pm0.013 $ \\
& & tox21 & graph & gin & max & $ 0.733\pm0.016  $ & $ 0.214\pm0.010  $ & $ 0.480\pm0.001 $ \\
\cline{3-9}
& & toxcast & SMILES & molt5 & avg & $ \textbf{0.673}\pm0.003  $ & $ \textbf{0.336}\pm0.017  $ & $ 0.465\pm0.007 $ \\
& & toxcast & InChI & t5 & avg & $ \textbf{0.673}\pm0.018  $ & $ 0.322\pm0.027  $ & $ \textbf{0.471}\pm0.001 $ \\
& & toxcast & SELFIES & molt5 & avg & $ 0.663\pm0.014  $ & $ 0.319\pm0.014  $ & $ 0.469\pm0.004 $ \\
& & toxcast & graph & gcn & avg & $ 0.662\pm0.033  $ & $ 0.325\pm0.030  $ & $ 0.464\pm0.007 $ \\
& & toxcast & SMILES & t511 & avg & $ 0.657\pm0.038  $ & $ 0.318\pm0.029  $ & $ 0.464\pm0.009 $ \\
\hline
\end{tabular}%
}
\label{tab:experimental_results_MPP_classification}
\end{table}

\begin{table}[ht!]
\centering
\caption*{\textbf{Supplementary Table 9:} Experimental results for regression tasks}
\resizebox{0.97\textwidth}{!}{%
\begin{tabular}{l|l|lllllll}
\hline
\textbf{Task} & \textbf{Data Spliting} & \textbf{Dataset} & \textbf{Input} & \textbf{Encoder}& \textbf{Pooling}  & \textbf{mse $\downarrow$} & \textbf{rmse $\downarrow$} & \textbf{mae $\downarrow$} \\
\hline
\multirow{30}{*}{\shortstack{\textbf{property prediction} \\ (regression)}} &
\multirow{15}{*}{scaffold} 
& esol & SMILES & t5 & avg & $  \textbf{1.573} \pm 0.139 $ & $  1.253 \pm 0.056 $ & $  \textbf{0.972} \pm 0.027$ \\
& & esol & SMILES & molt5 & avg & $  1.593 \pm 0.408 $ & $  \textbf{1.252} \pm 0.158 $ & $  0.974 \pm 0.102$ \\
& & esol & InChI & molt5 & avg & $  1.739 \pm 0.089 $ & $  1.318 \pm 0.034 $ & $  0.965 \pm 0.038$ \\
& & esol & graph & gat & avg & $  1.750 \pm 0.173 $ & $  1.321 \pm 0.066 $ & $  0.983 \pm 0.057$ \\
& & esol & SMILES & t511 & avg & $  1.910 \pm 0.533 $ & $  1.369 \pm 0.192 $ & $  1.039 \pm 0.125$ \\
\cline{3-9}
& & freesolv & graph & gcn & max & $  \textbf{9.701} \pm 1.892 $ & $  \textbf{3.101} \pm 0.295 $ & $  \textbf{2.254} \pm 0.225$ \\
& & freesolv & SELFIES & molt5 & avg & $  9.858 \pm 0.294 $ & $  3.139 \pm 0.047 $ & $  2.391 \pm 0.007$ \\
& & freesolv & graph & gat & avg & $  10.429 \pm 2.184 $ & $  3.212 \pm 0.332 $ & $  2.608 \pm 0.267$ \\
& & freesolv & SELFIES & t5 & avg & $  11.689 \pm 0.942 $ & $  3.416 \pm 0.140 $ & $  2.451 \pm 0.113$ \\
& & freesolv & SELFIES & t511 & avg & $  11.740 \pm 3.591 $ & $  3.389 \pm 0.504 $ & $  2.637 \pm 0.411$ \\
\cline{3-9}
& & lipophilicity & graph & gin & avg & $  \textbf{0.676} \pm 0.067 $ & $  \textbf{0.821} \pm 0.040 $ & $  \textbf{0.656} \pm 0.043$ \\
& & lipophilicity & graph & gcn & avg & $  0.735 \pm 0.006 $ & $  0.857 \pm 0.003 $ & $  0.684 \pm 0.007$ \\
& & lipophilicity & graph & gcn & max & $  0.770 \pm 0.093 $ & $  0.876 \pm 0.052 $ & $  0.699 \pm 0.046$ \\
& & lipophilicity & graph & gat & avg & $  0.855 \pm 0.067 $ & $  0.924 \pm 0.036 $ & $  0.736 \pm 0.025$ \\
& & lipophilicity & SMILES & molt5 & max & $  0.890 \pm 0.147 $ & $  0.940 \pm 0.077 $ & $  0.762 \pm 0.064$ \\
\cline{2-9}
&\multirow{15}{*}{random} 
&  esol & InChI & t511 & avg & $  \textbf{0.967} \pm 0.071 $ & $  \textbf{0.983} \pm 0.036 $ & $ \textbf{0.718} \pm 0.053$ \\
& & esol & SMILES & molt5 & avg & $  1.035 \pm 0.292 $ & $  1.007 \pm 0.142 $ & $  0.779 \pm 0.130$ \\
& & esol & SMILES & t511 & avg & $  1.201 \pm 0.306 $ & $  1.087 \pm 0.138 $ & $  0.831 \pm 0.126$ \\
& & esol & InChI & molt5 & avg & $  1.234 \pm 0.157 $ & $  1.109 \pm 0.070 $ & $  0.794 \pm 0.037$ \\
& & esol & graph & gcn & max & $  1.273 \pm 0.316 $ & $  1.120 \pm 0.136 $ & $  0.841 \pm 0.103$ \\
\cline{3-9}
& & freesolv & graph & gcn & max & $  \textbf{5.049} \pm 0.493 $ & $  \textbf{2.244} \pm 0.112 $ & $  \textbf{1.733} \pm 0.031$ \\
& & freesolv & graph & gat & max & $  5.175 \pm 1.119 $ & $  2.261 \pm 0.251 $ & $  1.818 \pm 0.226$ \\
& & freesolv & graph & gat & avg & $  5.536 \pm 0.275 $ & $  2.352 \pm 0.058 $ & $  1.706 \pm 0.059$ \\
& & freesolv & graph & gcn & avg & $  6.064 \pm 1.485 $ & $  2.443 \pm 0.306 $ & $  1.829 \pm 0.218$ \\
& & freesolv & InChI & t5 & avg & $  7.343 \pm 1.255 $ & $  2.700 \pm 0.232 $ & $  1.920 \pm 0.179$ \\
\cline{3-9}
& & lipophilicity & graph & gin & avg & $  \textbf{0.690} \pm 0.099 $ & $  \textbf{0.828} \pm 0.061 $ & $  \textbf{0.650} \pm 0.043$ \\
& & lipophilicity & graph & gat & avg & $  0.717 \pm 0.019 $ & $  0.846 \pm 0.011 $ & $  0.676 \pm 0.008$ \\
& & lipophilicity & graph & gcn & avg & $  0.785 \pm 0.077 $ & $  0.885 \pm 0.043 $ & $  0.693 \pm 0.045$ \\
& & lipophilicity & graph & gcn & max & $  0.795 \pm 0.132 $ & $  0.888 \pm 0.075 $ & $  0.704 \pm 0.080$ \\
& & lipophilicity & SMILES & molt5 & avg & $  0.887 \pm 0.221 $ & $  0.934 \pm 0.121 $ & $  0.740 \pm 0.100$ \\

\hline
\end{tabular}%
}
\label{tab:experimental_results_MPP_regression}
\end{table}
The scaffold data splitting method inherently increases the complexity of predictions.  Unlike molecular retrieval tasks, where the focus is on embedding tasks with textual targets, leading to a better performance of language models over graph encoders.
Here, graph encoders generally outperform language models in most cases.

\begin{table}[htbp]
\centering
\caption*{\textbf{Supplementary Table 10:} Multi-modal fusion performance for molecular property predictions.
This table displays the ROC\_AUC results for various molecular property prediction classification tasks.
``Add" denotes vector addition, ``Weight\_add" adaptive weighted addition, ``Concat" concatenated encoding with pooling, and ``Concat\_attention" concatenated encoding with attention before pooling.
The ``SMILES" and ``graph2d" columns indicate input modalities, with ``1" for inclusion and ``0" for exclusion.
The columns BBBP, Tox21, ToxCast, SIDER, ClinTox, and BACE represent datasets.
}
\resizebox{0.97\textwidth}{!}{%
\begin{tabular}{@{}l|l|cc|ccccccl@{}}
\toprule
Method                      & Model            & SMILES & graph2d & BBBP  & Tox21 & ToxCast & SIDER & ClinTox & BACE  & Avg   \\ \midrule
\textemdash                & SciBERT         & 1      & 0       & 0.553 & 0.508 & 0.524   & 0.512 & 0.445   & 0.438 & 0.497 \\
\textemdash                & GIN             & 0      & 1       & 0.640 & 0.701 & 0.593   & 0.504 & 0.504   & 0.648 & 0.598 \\
Add                        & SciBERT, GIN    & 1      & 1       & 0.651 & 0.696 & 0.603   & 0.581 & 0.503   & 0.559 & 0.599 \\
Weight\_add                & SciBERT, GIN    & 1      & 1       & 0.633 & 0.717 & 0.602   & 0.522 & 0.527   & 0.802 & 0.634 \\
Concat                     & SciBERT, GIN    & 1      & 1       & 0.642 & 0.713 & 0.603   & 0.582 & 0.627   & 0.672 & 0.640 \\
Concat\_attention          & SciBERT, GIN    & 1      & 1       & 0.620 & 0.582 & 0.557   & 0.489 & 0.676   & 0.630 & 0.592 \\
\hline
\textemdash                & MolT5           & 1      & 0       & 0.638 & 0.736 & 0.556   & 0.526 & 0.909   & 0.686 & 0.675 \\
Add                        & MolT5, GIN      & 1      & 1       & 0.673 & 0.724 & 0.614   & 0.559 & 0.844   & 0.541 & 0.659 \\
Weight\_add                & MolT5, GIN      & 1      & 1       & 0.634 & 0.716 & 0.577   & 0.556 & 0.886   & 0.751 & 0.687 \\
Concat                     & MolT5, GIN      & 1      & 1       & 0.643 & 0.702 & 0.578   & 0.570 & 0.731   & 0.600 & 0.637 \\
Concat\_attention          & MolT5, GIN      & 1      & 1       & 0.641 & 0.680 & 0.555   & 0.520 & 0.834   & 0.536 & 0.628 \\
\hline
\textemdash                & BioT5           & 1      & 0       & 0.681 & 0.702 & 0.598   & 0.557 & 0.890   & 0.699 & 0.688 \\
Add                        & BioT5, GIN      & 1      & 1       & 0.693 & 0.717 & 0.603   & 0.578 & 0.870   & 0.748 & 0.702 \\
Weight\_add                & BioT5, GIN      & 1      & 1       & 0.721 & 0.713 & 0.589   & 0.583 & 0.875   & 0.701 & 0.697 \\
Concat                     & BioT5, GIN      & 1      & 1       & 0.706 & 0.705 & 0.599   & 0.557 & 0.814   & 0.720 & 0.683 \\
Concat\_attention          & BioT5, GIN      & 1      & 1       & 0.644 & 0.687 & 0.598   & 0.566 & 0.903   & 0.552 & 0.658 \\
\hline
\textemdash                & KV-PLM          & 1      & 0       & 0.720 & 0.700 & 0.550   & 0.598 & 0.892   & 0.785 & 0.708 \\
\hline
Contrastive learning       & MoMu-SciBERT    & 0      & 1       & 0.705 & 0.756 & 0.634   & 0.605 & 0.799   & 0.767 & 0.711 \\
                           & MoMu-KV-PLM     & 0      & 1       & 0.701 & 0.756 & 0.630   & 0.604 & 0.774   & 0.771 & 0.706 \\
                           & MoleculeSTM     & 1      & 0       & 0.708 & 0.757 & 0.652   & 0.637 & 0.866   & 0.820 & 0.740 \\
                           & MoleculeSTM     & 0      & 1       & 0.700 & 0.769 & 0.651   & 0.610 & 0.925   & 0.808 & 0.744 \\
\hline
Contrastive learning, & MolFM & 0 & 1 & 0.722 & 0.766 & 0.642 & 0.632 & 0.786 & 0.826 & 0.729 \\
cross attention & MolFM           & 1      & 1       & 0.729 & 0.772 & 0.644   & 0.642 & 0.797   & 0.839 & 0.737 \\
                           & MolCA           & 1      & 0       & 0.708 & 0.760 & 0.562   & 0.611 & 0.890   & 0.793 & 0.721 \\
                           & MolCA           & 1      & 1       & 0.700 & 0.772 & 0.645   & 0.630 & 0.895   & 0.798 & 0.740 \\
                           & GIT-Mol         & 1      & 0       & 0.719 & 0.739 & 0.621   & 0.601 & 0.835   & 0.684 & 0.700 \\
                           & GIT-Mol         & 0      & 1       & 0.711 & 0.754 & 0.653   & 0.582 & 0.789   & 0.658 & 0.691 \\
                           & GIT-Mol         & 1      & 1       & 0.739 & 0.759 & 0.668   & 0.634 & 0.883   & 0.811 & 0.749 \\ \bottomrule
\end{tabular}%
}
\end{table}

\begin{table}[htbp]
\centering
\caption*{\textbf{Supplementary Table 11:} Multi-modal fusion performance for molecule captioning tasks.
This table presents the performance of six textual similarity metrics (BLEU-2, BLEU-4, ROUGE-1, ROUGE-2, ROUGE-L, and Meteor) across two common datasets, ChEBI-20 and PubChem324k, for molecule captioning tasks.
The ``Method" column includes ``CL" for contrastive learning and ``CA" for cross attention.
The ``SMILES" and ``graph2d" columns indicate input modalities, with ``1" for inclusion and ``0" for exclusion.}
\resizebox{\textwidth}{!}{%
\begin{tabular}{@{}l|l|c|c|c|cccccl@{}}
\toprule
\textbf{Benchmark} & \textbf{Method}                      & \textbf{Model}        & \textbf{SMILES} & \textbf{graph2d} & \textbf{BLEU-2} & \textbf{BLEU-4} & \textbf{ROUGE-1} & \textbf{ROUGE-2} & \textbf{ROUGE-L} & \textbf{Meteor} \\ \midrule
\multirow{8}{*}{ChEBI-20} 
    & \multirow{3}{*}{\textemdash} & RNN          & 1      & 0       & 0.251  & 0.176  & 0.450   & 0.278   & 0.394   & 0.363        \\
    &                               & Transformer  & 1      & 0       & 0.061  & 0.027  & 0.204   & 0.087   & 0.186   & 0.114        \\
    &                               & MolT5        & 1      & 0       & 0.540  & 0.457  & 0.634   & 0.485   & 0.578   & 0.569        \\ \cline{2-11}
    & \multirow{3}{*}{CL}           & MoMu         & 0      & 1       & 0.549  & 0.462  & 0.000   & 0.000   & 0.575   & 0.576        \\
    &                               & InstructMol  & 0      & 1       & 0.466  & 0.365  & 0.547   & 0.365   & 0.479   & 0.491        \\
    &                               & InstructMol  & 1      & 1       & 0.475  & 0.371  & 0.566   & 0.394   & 0.502   & 0.509        \\ \cline{2-11}
    & \multirow{2}{*}{CL+CA}        & MolFM        & 1      & 1       & 0.585  & 0.498  & 0.653   & 0.508   & 0.594   & 0.607        \\
    &                               & MolCA        & 1      & 1       & 0.620  & 0.531  & 0.681   & 0.537   & 0.618   & 0.651        \\ \midrule
\multirow{7}{*}{PubChem324k} 
    & \multirow{1}{*}{\textemdash} & MolT5        & 1      & 0       & 0.301  & 0.209  & 0.403   & 0.251   & 0.338   & 0.356        \\ \cline{2-11}
    & \multirow{1}{*}{CL}           & MoMu         & 0      & 1       & 0.302  & 0.215  & 0.405   & 0.251   & 0.344   & 0.342        \\ \cline{2-11}
    & \multirow{3}{*}{CL+CA}        & MolCA        & 1      & 0       & 0.346  & 0.269  & 0.463   & 0.323   & 0.415   & 0.411        \\
    &                               & MolCA        & 0      & 1       & 0.345  & 0.262  & 0.464   & 0.316   & 0.412   & 0.409        \\
    &                               & MolCA        & 1      & 1       & 0.387  & 0.303  & 0.502   & 0.359   & 0.445   & 0.456        \\ \cline{2-11}
    & \multirow{2}{*}{}             & GIT-Mol      & 1      & 0       & 0.264  & 0.176  & 0.477   & 0.374   & 0.451   & 0.430        \\
    &                               & GIT-Mol      & 0      & 1       & 0.290  & 0.210  & 0.540   & 0.445   & 0.512   & 0.491        \\
    &                               & GIT-Mol      & 1      & 1       & 0.352  & 0.263  & 0.575   & 0.485   & 0.560   & 0.533        \\ \bottomrule
\end{tabular}%
}
\end{table}

\clearpage

\section{Comparative Analysis of Interpretation Approaches}

As indicated in Supplementary Table 12, we review commonly used interpretability methods in the domain of large molecular models.
Most research utilizes approaches based on sample embedding visualization or inductive methods from case studies of text generation.
However, embedding visualization primarily reflects changes in molecular representations, explaining how models understand and align data, but it does not necessarily offer interpretability to humans.
Methods based on case studies require extensive time to extract insights, which is time-consuming and lacks statistical validation.
Token frequency analysis is another common approach, yet selecting the appropriate tokens poses a significant challenge.
Conventional methods like the Standard Deviation Method and the Quartile Method often result in the selection of `general high-frequency token mappings.'
In contrast, our Localized Feature Filtering Method is capable of identifying `particular high-frequency token mappings,' providing more targeted and meaningful interpretative insights.

\begin{table}[h!]
\centering
\caption*{\textbf{Supplementary Table 12:} Comparative analysis of interpretation approaches}
\label{tab:Comparative_methods}
\resizebox{\textwidth}{!}{%
\begin{tabular}{@{}l|l|l|p{5cm}@{}}
\toprule
\textbf{Task} & \textbf{Approach} & \textbf{Mode} & \textbf{Research} \\
\midrule
\multirow{2}{*}{Representation} & Embedding Visualization & Samples Induction & SMPT 
 \cite{li2024pre}, GIT-Mol \cite{liu2023git}, MV-Mol \cite{luo2024learning}, KV-PLM \cite{zeng2022deep}, MoMu \cite{su2022molecular}, Polymer\_Walk \cite{sunrepresenting} \\
\cline{2-4}
 & Rule-based Analysis & Inference & Polymer\_Walk \cite{sunrepresenting} \\
\midrule
\multirow{2}{*}{Text Generation} & Case Study Analysis & Case Study Induction & ChatMol \cite{zeng2023interactive}, GIT-Mol, MolReGPT \cite{li2023empowering}, MolFM \cite{luo2023molfm}, KV-PLM, Atomas \cite{zhang2024atomas} \\
\cline{2-4}
 & Atom Attention Visualization & Case Study Induction & MolFM \\
\midrule
Molecule Generation & Substructure Attention Visualization & Case Study Induction & MOLGEN \cite{fang2024domainagnostic} \\
\midrule
\multirow{3}{*}{Text/Molecule Generation} & \multirow{3}{*}{Tokens Frequency Analysis} 
 &  Standard Deviation Method & -- \\
 & & Quartile Method  &  --\\
 & & \textbf{Localized Feature Filtering Method} & Ours \\
\bottomrule
\end{tabular}}
\end{table}

\clearpage


\end{document}